\documentclass[sigconf]{acmart}
\usepackage{rotating}
\usepackage{hyperref}

\AtBeginDocument{%
  \providecommand\BibTeX{{%
    \normalfont B\kern-0.5em{\scshape i\kern-0.25em b}\kern-0.8em\TeX}}}

\usepackage{multicol}
\usepackage{multirow}
\usepackage{array}
\usepackage{cleveref}
\usepackage{subcaption}
\usepackage{multicol}
\usepackage{multirow}
\usepackage{enumitem}
\usepackage{soul}

\usepackage{booktabs,array}
\newcommand{\q}[1]{\emph{``#1''}}
\usepackage{multirow}
\usepackage{subcaption}
\definecolor{ForestGreen}{RGB}{34,139,34}

\usepackage{pifont}

\usepackage{pifont}

\RequirePackage{color}\definecolor{RED}{rgb}{1,0,0}\definecolor{BLUE}{rgb}{0,0,1} 
\RequirePackage{listings} 
\RequirePackage{color} 

\copyrightyear{2026}
\acmYear{2026}
\setcopyright{cc}
\setcctype{by}
\acmConference[CHI '26]{Proceedings of the 2026 CHI Conference on Human Factors in Computing Systems}{April 13--17, 2026}{Barcelona, Spain}
\acmBooktitle{Proceedings of the 2026 CHI Conference on Human Factors in Computing Systems (CHI '26), April 13--17, 2026, Barcelona, Spain}
\acmPrice{}
\acmDOI{10.1145/3772318.3791276}
\acmISBN{979-8-4007-2278-3/2026/04}

\begin{document}

\copyrightyear{2026}
\acmYear{2026}
\setcopyright{cc}
\setcctype{by}
\acmConference[CHI '26]{Proceedings of the 2026 CHI Conference on Human Factors in Computing Systems}{April 13--17, 2026}{Barcelona, Spain}
\acmBooktitle{Proceedings of the 2026 CHI Conference on Human Factors in Computing Systems (CHI '26), April 13--17, 2026, Barcelona, Spain}
\acmPrice{}
\acmDOI{10.1145/3772318.3791276}
\acmISBN{979-8-4007-2278-3/2026/04}

\title{Can Good Writing Be Generative? Expert-Level AI Writing Emerges through Fine-Tuning on High-Quality Books}

\author{Tuhin Chakrabarty}
\affiliation{%
  \institution{Stony Brook University}
  \city{Stony Brook}
  \state{NY}
  \country{USA}}
  \email{tchakrabarty@cs.stonybrook.edu}

\author{Paramveer S. Dhillon}
\affiliation{%
  \institution{University of Michigan}
  \city{Ann Arbor}
  \state{MI}
  \country{USA}}
  \email{dhillonp@umich.edu}

\begin{abstract}
Creative writing has long been considered a uniquely human endeavor, requiring voice and style that machines could not replicate. This assumption is challenged by Generative AI that can emulate thousands of author styles in seconds with negligible marginal labor. To understand this better, we conducted a behavioral experiment where 28 MFA writers (experts) competed against three LLMs in emulating 50 critically acclaimed authors. Based on blind pairwise comparisons by 28 expert judges and 131 lay judges, we find that experts preferred human writing in 82.7\% of cases under the in-context prompting condition but this reversed to 62\% preference for AI after fine-tuning on authors' complete works. Lay judges, however, consistently preferred AI writing.  Debrief interviews with expert writers revealed that their preference for AI writing triggered an identity crisis, eroding aesthetic confidence and questioning what constitutes ``good writing.'' These findings challenge discourse about AI's creative limitations and raise fundamental questions about the future of creative labor.
\end{abstract}




\maketitle

\section{Introduction}
In his book ``The Program Era" \cite{mcgurl2011program}, eminent literary critic Mark McGurl discussed how "\textit{The rise of the creative-writing program stands as the most important event in postwar American literary history.}" In these programs, "Show, don't tell," which was a slogan in the nineteen-forties and fifties, shifted to an effectively opposite mantra, "Find your voice," that took over in the nineteen-sixties and seventies. McGurl observed that these guiding principles reflect evolving cultural beliefs—about identity, labor, gender, class, and notions of what counts as \textit{good writing}—and that they have had a huge effect on the stories and novels that American writers have produced \citep{menand2009show}. McGurl's observation about what constitutes "good writing" takes on new significance in the age of generative artificial intelligence. The same qualities that creative writing programs cultivated—coherence, voice, narrative structure—are precisely what make their literary output valuable training data for LLMs \cite{gero2025creative}. Models trained on well-edited books and articles tend to generate more coherent and accurate outputs, something that's crucial for establishing LLMs as intelligent systems. 

However extensive social experimentation with Generative AI has invited criticism on social media and in the popular news platforms that its writing has a disembodied ``robovoice". Recent work from \citet{chakrabarty2025can} shows how text generated from widely used LLMs are often rife with clichés, purple prose, poor sentence structure, and unnecessary exposition. AI writing often veers towards homogenization \cite{doshi2024generative,anderson2024homogenization} that is directly in contrast with what creative writing programs have taught. To counter this, some users and practitioners now prompt or train language models\footnote{\url{https://community.openai.com/t/fine-tuning-the-model-for-writing-via-famous-author-style/512322}} to emulate a specific writer's style/voice \cite{Tangermann2025}. This was also acknowledged by the US Copyright Office in their Generative AI Training report Part 1, where many commenters raised concerns about AI outputs imitating a creator’s style, which copyright does not protect as a separate element. Based on the U.S. Bureau of Labor Statistics (May 2023, national estimates), creative writing constitutes almost 50\% of writing jobs that face significant risks from Generative AI-based automation.\footnote{\url{https://lithub.com/against-ai-an-open-letter-from-writers-to-publishers/}}

\begin{figure*}[t]
    \centering
    \begin{subfigure}{\textwidth}
        \centering
        \fbox{\includegraphics[width=0.98\linewidth]{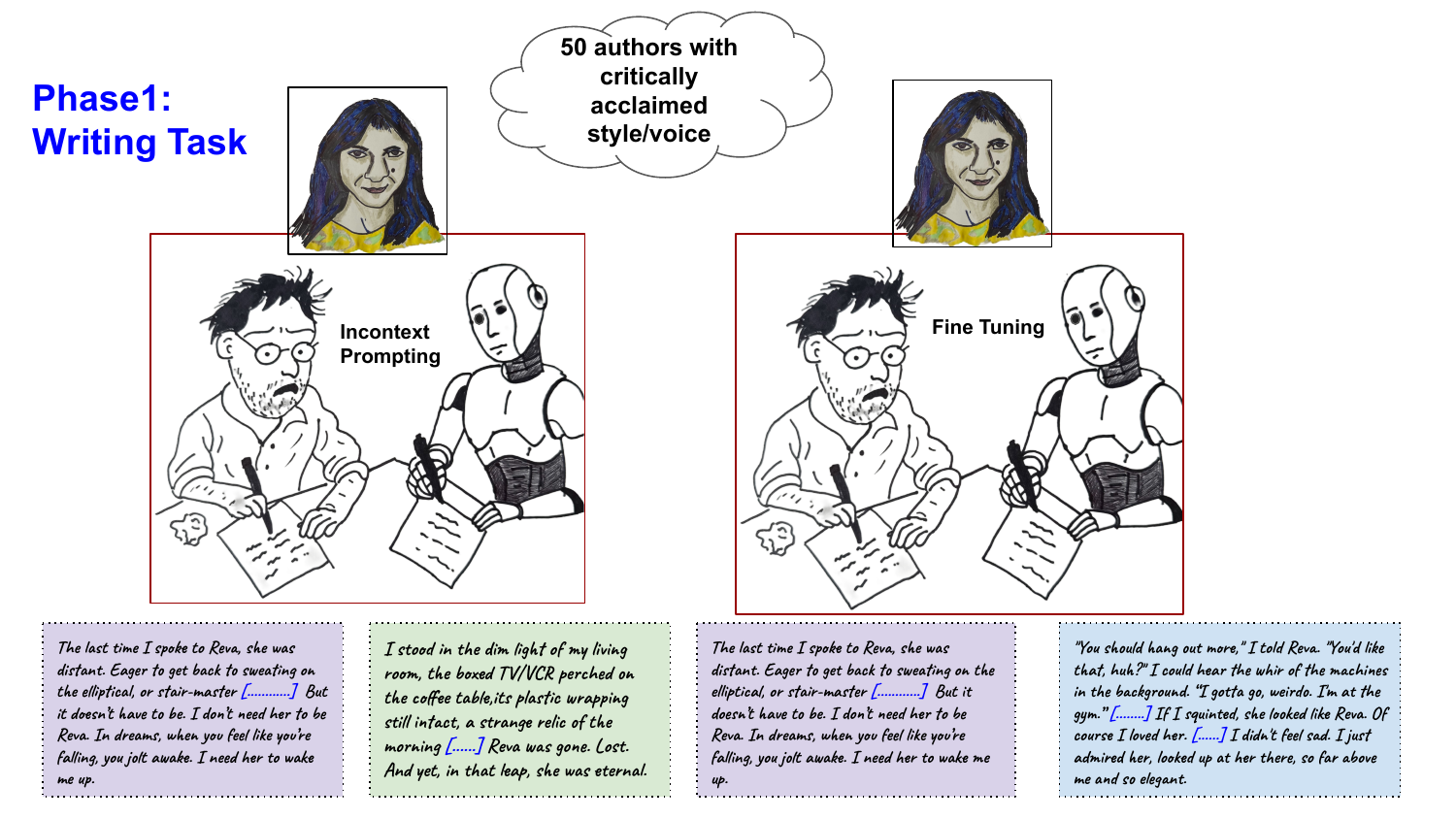}} 
        \vspace{-1ex}
    \end{subfigure}  
    \begin{minipage}[t]{0.49\textwidth}
        \centering
        \includegraphics[width=\linewidth]{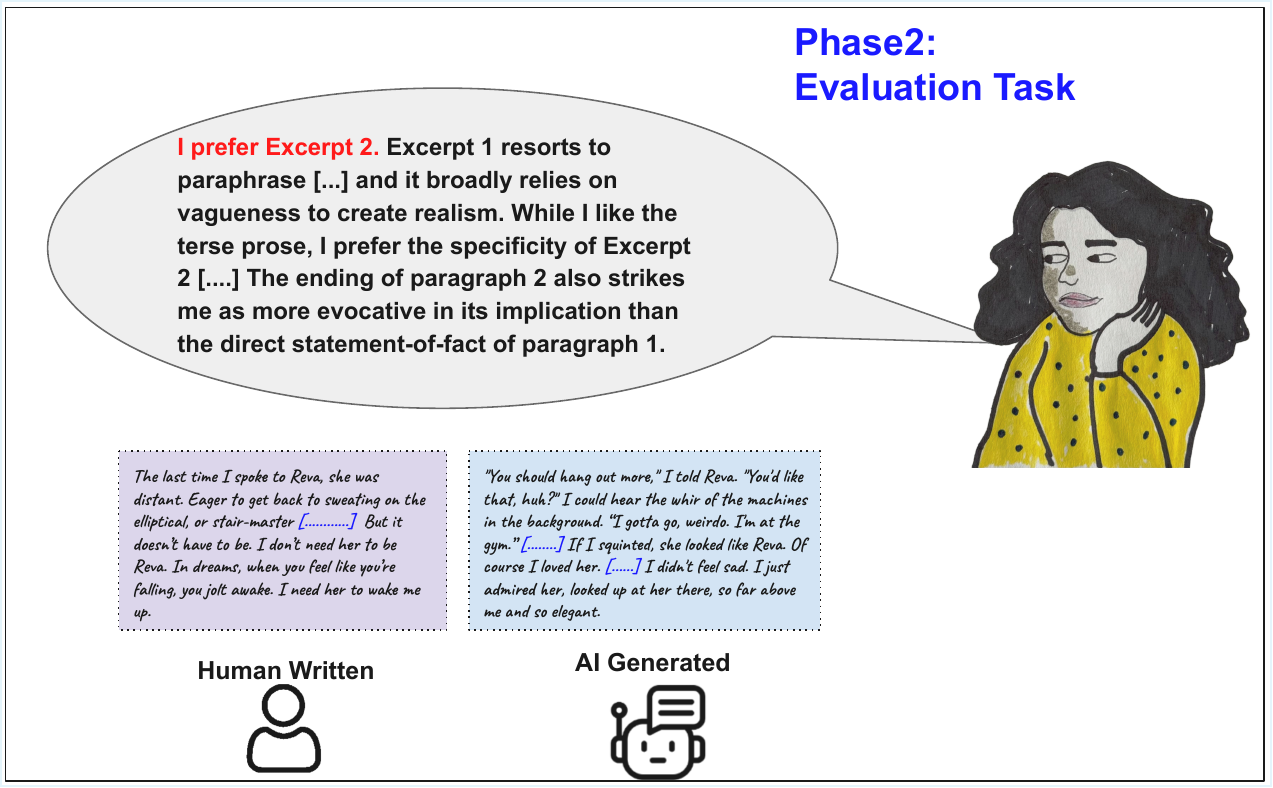}
    \end{minipage}
    \hfill 
    \begin{minipage}[t]{0.49\textwidth}
        \centering
         \includegraphics[width=\linewidth, height=7cm,keepaspectratio]{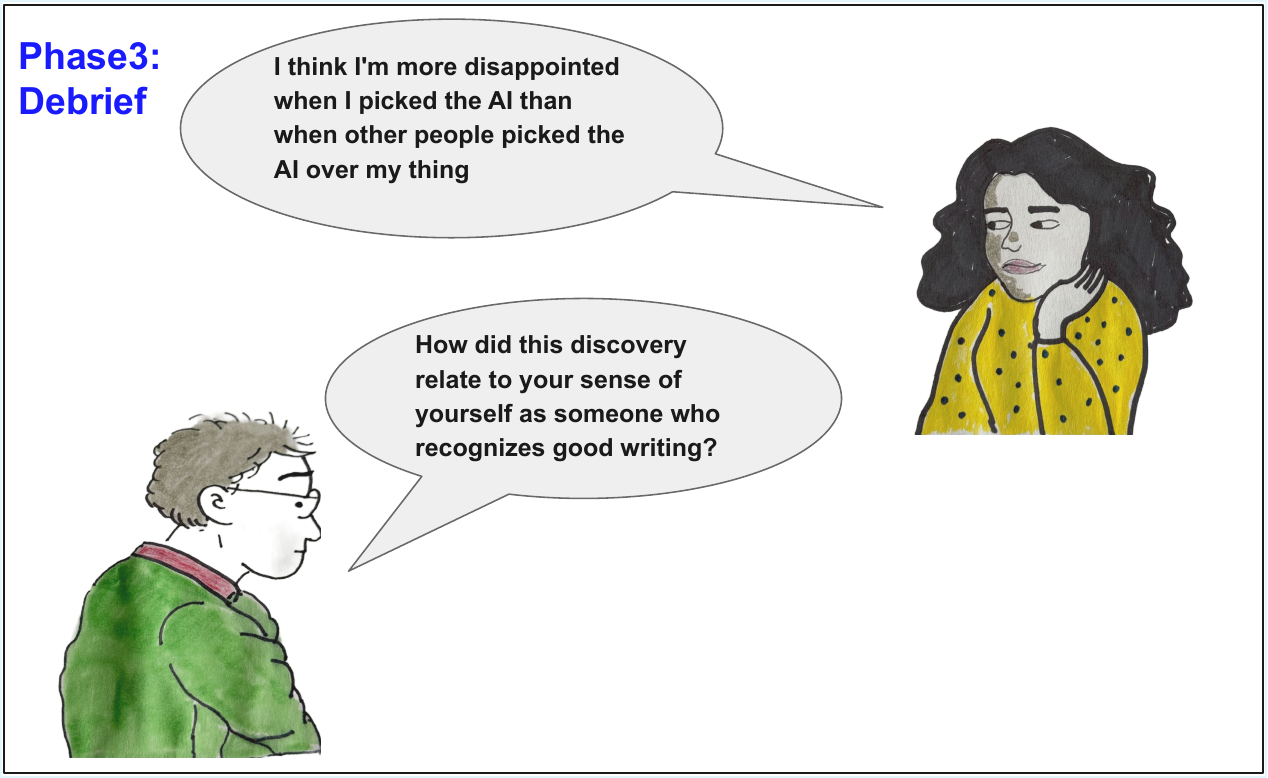}
    \end{minipage}
    \caption{Three-phase study design showing (Phase 1) writing task with In-context prompting and fine-tuning approaches using authors with critically acclaimed style/voice. In the figure both writers and AI are trying to write a specific excerpt from \textit{My Year of Rest and Relaxation} in the style/voice of Ottessa Moshfegh (Phase 2) Human-written vs AI-generated text evaluation task with rationales, and (Phase 3) Debrief session exploring sentiments and sensemaking when writers prefer AI over human writing. \tiny{Figure illustrated by the first author of the paper}}
    \Description{Three-panel illustration of study design: Phase 1 shows a human writer and AI system both attempting to write in the style of author Ottessa Moshfegh, with the AI using either in-context prompting or fine-tuning on the author's works. Phase 2 shows an evaluator comparing human-written and AI-generated text side-by-side, providing preference judgments and written rationales. Phase 3 depicts a debrief interview session where participants discuss their reactions to learning AI was preferred over human writing.}
    \label{fig:study_findings}
    \vspace{-2ex}
\end{figure*}

To understand how much ``good writing" comes from having a distinct style/voice we conducted a behavioral study where 28 expert human writers were pitted against three LLMs (See Figure \ref{fig:study_findings}). We first preselected a group of 50 authors representing diverse cultural backgrounds and age groups, who are known for their critically acclaimed literary style/voice. We then assigned both human expert writers and LLMs the same task: to write a 200-450 word excerpt emulating the style and voice of a selected author, based on a writing prompt. For LLMs we included two conditions: (1) In-context Prompting, where models received the same prompt as human expert writers and (2) Fine-tuning, where models were additionally trained on each author’s complete oeuvre. There are several ways to get an LLM to emulate the style/voice of an author. In-context prompting puts representative examples in a prompt, whereas Fine-tuning actually modifies the model through training on author's books and is more resource intensive. Following this, we recruited the same 28 writers along with 131 lay human participants from Prolific who acted as judges to evaluate these excerpts in a blind evaluation across two different axes i) Writing Quality ii) Stylistic Fidelity. We refer to writers as expert judges. Through planned pairwise contrasts between $<Human\_written, AI\_generated>$ excerpts, both expert and lay judges provided their preferences as well as 2-5 sentence rationales grounded in snippets from these excerpts, explaining their preference.

\begin{itemize}
    \item \textbf{RQ1}: What line of reasoning do experts and lay judges use to justify what qualifies as good writing?
    \item \textbf{RQ2}: After learning a preferred excerpt was written by AI, how do writers reconcile with others as well as their own judgments?
    \item \textbf{RQ3}: How does the discovery of choosing AI over human writing affect writers’ sense of professional identity as well as their fundamental understanding of what constitutes “good writing."
    \item \textbf{RQ4}: How do writers interpret AI's advanced capabilities in terms of what they mean for creative writing programs and their writing career—and what changes do they suggest for educational institutions, publishing houses, and literary platforms?\end{itemize}

In addressing RQ1, we analyze the rationales collected alongside each blind preference. We find that lay judges primarily focus on surface-level qualities like flow, organization, and clarity, often citing whether writing is ``easier to follow" or has emotional impact without analyzing the mechanisms behind it. Experts, in contrast, provide more detailed analysis focusing on narrative voice, character interiority, specific literary devices like imagery and syntax, and how technical elements serve the overall meaning and authenticity of the work. To answer RQ2–RQ4 we draw on retrospective semi-structured interviews with 21 writers who participated in the study. After learning their preferred response was AI-generated, we find writers reconcile with their previous judgments in five key ways. These involve {\it \textbf{Criteria Reframing}} (questioning their evaluation criteria around what constitutes ``good writing,"), {\it \textbf{Process Attribution}} (contrasting human versus AI creative processes), {\it \textbf{Technical Sensemaking}} (attempting to rationalize AI's capabilities through technical understanding), {\it \textbf{Expectation Violation}} (experiencing psychological shock when expectations are violated) and {\it \textbf{Capability Reassessment}} (fundamentally reassessing their beliefs about AI's creative limitations) (RQ2). The discovery of choosing AI over human writing leads to an {\it \textbf{Erosion of Aesthetic Confidence}} in writers with many expressing that they can no longer confidently distinguish high-quality AI writing from human writing. This in turn leads to an {\it \textbf{Identity Crisis Around Expertise}} where writers question their expertise  or trust their professional abilities. Additionally this destabilization leads writers to {\it \textbf{Redefine Writing's Purpose}} away from aesthetic quality toward process and intention (RQ3). Last but not least writers view AI's advanced capabilities as a threat to their livelihoods especially given the nature of publishing and market dynamics. They emphasize the need for i) creative writing programs to prioritize truly exceptional and unique styles ii) for publishing houses and online platforms to require AI disclosure and strengthen copyright protections and iii) for the literary community to provide greater solidarity through unions and advocacy to prevent the commercialization of AI-generated content (RQ4).

In the discussion, we focus on prevalent discourse around how AI emulating style/voice is different from writers being influenced by the style of other writers. We also discuss how our research adds to the current debate on Generative AI and Fair use while providing conflicting evidence on labor market dilution. Finally, we discuss pathways around regulating hidden AI authorship and what it means for future of creative work. All the writing prompts and preference data along with rationales can be found in the provided link. \footnote{\url{https://github.com/tuhinjubcse/GoodWritingBeGenerative}}.

\section{RELATED WORK}
\subsection{The role of AI in Writing}
Large language models (LLMs) have proven themselves as potent writing support tools due to their ability to generate coherent text, suggest improvements, and help overcome writer's block instantly. They have significantly changed how people write across various fields \cite{lee2024design}, with studies showing that 10-24\% of content in areas like consumer complaints, business communications, job listings, and UN press statements now involves LLM assistance as of late 2024 \cite{liang2025widespread}. Beyond standard professional writing, LLMs have found applications in both scientific research \citep{kobak2025delving, liang2024mapping, liang2025quantifying, gero2022sparks} and creative endeavors, including collaborative fiction writing and other artistic text generation \citep{10.1145/3635636.3656201, calderwood2025phraselette, mirowski2024robot, ippolito2022creative,yuan2022wordcraft,mirowski2023cowriting}. In prior work \citet{gero2023social} highlight the social dynamics that govern writers' collaboration with outside entities, and find that the types of support desired can be aligned with the updated cognitive process model of writing. However, while writing support promises productivity, it also leads to a decline in ownership and authenticity \cite{draxler2024ai,zhang2026owns}. \citet{guo2025pen} find that writers are intentional about how they incorporate AI, making many deliberate decisions about when and how to engage AI based on their core values, such as authenticity and craftsmanship.
\citet{dhillon2024shaping} emphasized the need for personalized scaffolding mechanisms in AI-powered writing tools. On the contrary, \citet{hwang202580} find that while writers reacted positively to personalized AI writing tools, they believed the form of personalization needs to target writers' growth \cite{zhao2025making} and go beyond the phase of text production. In a similar vein \citet{10.1145/3715336.3735848} also argued for how AI writing assistants could be more effective by initiating identity reflection and connecting young adults with their support networks rather than assuming private, proactive writing. While these studies have examined AI's role as a collaborative writing tool and its impact on writer identity and ownership, they primarily focus on AI assistance during the writing process rather than AI's ability to independently generate high-quality text. Rather than examining AI as a supportive tool, we investigate what happens when AI becomes indistinguishable from or even preferred over expert human writers in real life.

\subsection{Writing as training data for Large Language Models}
Using writing as training data for large language models has long precedence in the field of artificial intelligence. GPT-3 \cite{brown2020language} one of the first large language models was trained on the infamous BooksCorpus\footnote{\url{https://en.wikipedia.org/wiki/BookCorpus}} in addition to Common Crawl and WebText which consisted of many books. The first Llama model \cite{touvron2023llama} states its book sources are Project Gutenberg\footnote{\url{https://www.gutenberg.org/}} and Books3\footnote{\url{https://www.wired.com/story/battle-over-books3/}}, totaling 177 GB of books data in its training. It's widely known that Meta's Llama3 and Anthropic's Claude models have been trained on Libgen. Recent investigation on LibGen from \citet{hansen_bartz_anthropic_2025} shows publication dates of books used to train LLMs span multiple decades rather than focusing on recent releases. For nonfiction, approximately 635,000 books were published before 2000, while about 1.3M were published after 2000. Fiction shows a different distribution, with only 66,000 books dated before 2000 and 680,000 dated after 2000. These books are represented by publishers from major houses alongside numerous smaller publishers. Leading fiction publishers include HarperCollins and Penguin Random House, while nonfiction is dominated by academic publishers like Springer, Oxford University Press, and Cambridge University Press. Such practices have led to outrage from the community of writers. In response to these contentious training data practices \citet{gero2025creative} interviewed writers about LLM training data usage. Through their interviews they find that writers express core principles around supporting creative chains and respecting human creativity, but recognize tensions with realistic constraints like lack of control and industry-scale impacts, ultimately showing more concern about power imbalances than the technology itself. Unlike prior work that examined use of books as a part of LLM training or writer attitudes toward their work being used as training data, our study tests whether AI can produce high-quality writing that readers prefer over text from expert human writers through controlled experiments. In addition to it we also understand how writers rationalize and feel about choosing AI writing to be better than human-written text in absence of disclosure.

\subsection{Generative AI and Collision with Copyright Law}
In the recent copyright case Bartz v. Anthropic \cite{usdcc_ndca_bartz_anthropic_2025}, Judge Alsup observed that Anthropic obtained approximately five million books through LibGen and an additional two million via Pirate Library Mirror (PiLiMi). Court documents from the Bartz v. Anthropic case further disclosed that Anthropic physically dismantled millions of printed books, converted them to digital formats through scanning, and discarded the physical copies, with the express purpose of using this material to train Claude. While the U.S. has yet to produce determinative case law on whether inputting works into a generative AI system's training data constitutes fair use, this ambiguity creates new challenges for the legal system and corresponding policies. \citet{samuelson2023generative} adds how Generative AI seems poised to have substantial impacts on the careers of professional writers and artists. During the 2023 WGA strike, for instance, uses of generative AI was one focus of negotiations with screenwriters being worried that these technologies will displace them or diminish their compensation. \citet{10.1093/jiplp/jpaf046} emphasizes how the effect of Generative AI copying impacts currently extant works by putting them in competition with AI-generated outputs. The copyrighted works serve as fodder for new productions for the same markets (romance novels; personalized streaming content) in which the copied works operate. Research in HCI has further explored stakeholder perspectives on these copyright and authorship questions. \cite{lima2025public} investigate 432 laypeople's perceptions and found that people believe AI-generated art requires creativity and effort but not skill, with participants most likely to attribute authorship and copyright to the AI users and original training data artists, while showing egocentric bias by rating their own AI creations higher, especially when money was at stake. On the contrary \cite{he2025contributions} found that knowledge workers consistently give AI systems less credit than human partners for equivalent contributions to co-created writing and editing work, while still viewing disclosure of AI involvement as important. \cite{kyi2025governance} interview creative workers across visual art, writing, and programming, examining their experiences of harm from generative AI—particularly around unconsented use of their work for training and makes recommendations for more ethical AI governance that addresses gaps between current practices and creators' needs for consent, compensation, and credit. Unlike prior work that looked at creators' attitudes toward their work being used as training data or stakeholder perspectives on copyright questions, our study focuses on whether AI can produce high-quality writing that leads to expert readers preferring it over text from professional human writers. Additionally, while previous work focused on ethical and legal questions around AI training practices, we explore how writers psychologically reconcile with and rationalize their own aesthetic preferences when they unknowingly choose AI-generated text over human writing.

\subsection{Future of Creative Labor}
Using Generative AI for Creative Work has its own share of ramifications. One of them being homogenization of ideas. In prior work \cite{wadinambiarachchi2024effects,doshi2024generative,anderson2024homogenization} have shown how using Generative AI in creative work reduces the collective diversity of the output. Illustrators and creators are threatened by Generative AI especially given its ability to extract and reproduce any artist's individual style. Understanding the results of style transfer as ``boundary objects,'' \citet{porquet2025copying}  analyze how they can simultaneously be considered unsuccessful by artists and poised to replace their work by others.  \citet{sobel2024elements} notes that ``style is a holistic attribute of a work, or a group of works, that comprises a constellation of expressive choices. These expressive choices might be unprotectable individually, but in combination, they may constitute protectable expression. Protectable style is not necessarily limited to expression in one discrete work". So in the future laws might need to protect a creator's style especially if multiple of their copyrighted works are ingested to create competing copies or even new style that combines existing styles. \citet{erickson2024ai} empirically investigated conditions of work and mechanisms by which human creative labor might be replaced or displaced by Al technology. Their findings suggest that in the firms analyzed, AI use did not result in replacement of human labor but rather a "redefinition of roles and skills" and a "complex process of 'invisibilisation' of human inputs within AI-made products.

\section{Methodology}
\subsection{Writing Task}
\subsubsection{Recruitment}

\begin{table*}[h]
\centering
\begin{tabular}{c||c|c|c|c|c}
\hline
\textbf{Writer ID} & \textbf{Qualification} & \textbf{Gender} & \textbf{Writer ID} & \textbf{Qualification} & \textbf{Gender} \\
\hline
W1 & MFA Candidate in Fiction & Male & W15 & MFA Candidate in Fiction & Female \\
\hline
W2 & MFA Candidate in Fiction & Male & W16 & MFA Candidate in Fiction & Female \\
\hline
W3 & MFA Candidate in Fiction & Male & W17 & MFA Candidate in Fiction & Female \\
\hline
W4 & MFA Candidate in Nonfiction & Female & W18 & MFA Candidate in Fiction & Female \\
\hline
W5 & MFA Candidate in Fiction & Male & W19 & MFA Candidate in Fiction & Male \\
\hline
W6 & MFA Candidate in Fiction & Female & W20 & MFA Candidate in Fiction & Female \\
\hline
W7 & MFA Candidate in Fiction & Female & W21 & MFA Candidate in Fiction & Female \\
\hline
W8 & MFA Candidate in Fiction & Male & W22 & MFA Candidate in Fiction & Female \\
\hline
W9 & MFA Candidate in Fiction & Female & W23 & MFA Candidate in Fiction & Female \\
\hline
W10 & MFA Candidate in Fiction & Female & W24 & MFA Candidate in Fiction & Male \\
\hline
W11 & MFA Candidate in Fiction & Female & W25 & MFA Candidate in Fiction & Female \\
\hline
W12 & MFA Candidate in Fiction & Male & W26 & MFA Candidate in Fiction & Female \\
\hline
W13 & MFA Candidate in Fiction & Male & W27 & PhD in English & Male \\
\hline
W14 & MFA Candidate in Poetry & Non-Binary & W28 & MFA Candidate in Fiction & Female \\

\end{tabular}
\vspace{1ex}
\caption{\label{writers}Details of writers recruited for our study}
\end{table*}

For our writing task, we recruited 28 writers. Twenty-seven of them completed or are currently pursuing their MFA degree in top writing programs such as the Helen Zell Writers' Program at the University of Michigan, Iowa Writers' Workshop, Program in Creative Writing at NYU, Creative Writing Program at BU, and the MFA in Writing at Columbia School of the Arts. Sixteen of our writers identified as female, while 11 identified as male and one as non-binary. Table \ref{writers} provides details about individual writers. Our writers ranged in age from 25 to 35 and were racially diverse, identifying as Hispanic, Black, Asian, and White. Our expert sample represents the emerging literary elite whose professional judgment shapes contemporary publishing. Among our recruited MFA candidates, several have since been awarded Stanford Stegner Fellowships (one of the most competitive fiction writing fellowships in the United States), received Rhodes Scholarships, Pushcart Prize, assumed editorial positions at prestigious literary magazines such as \textit{Joyland}, been awarded the Publishers Weekly Star Watch 2025 Honor and secured publishing contracts with major houses including W.W.~Norton and HarperCollins. Additionally almost each of the recruited writers have had their fiction published in prestigious literary magazines such as \textit{The Yale Review, The Greensboro Review, Southampton Review, }\\ \textit{The Missouri Review, Joyland Magazine, Iowa Review}. These accomplishments underscore that our experts represent not merely MFA candidates, but established and emerging voices whose aesthetic judgments carry professional authority in the literary marketplace. We acknowledge that all our recruited experts resided in the United States at the time of the study. This geographic restriction was necessary for administrative compliance, as U.S. tax regulations require declaration of compensation above certain thresholds, and the university's payment infrastructure could not accommodate international participants without Individual Taxpayer Identification Numbers (ITINs) or Social Security Numbers (SSNs).

\begin{table*}[!ht]
\small
\centering
\begin{tabular}{r l  r l  r l}
\toprule
\# & Author & \# & Author & \# & Author \\
\midrule
 1 & Alice Munro            & 19 & J.D. Salinger        & 37 & Philip Roth \\
 2 & Annie Ernaux (\checkmark)          & 20 & Jhumpa Lahiri (\checkmark)       & 38 & Rachel Cusk  (\checkmark)\\
 3 & Annie Proulx  (\checkmark)         & 21 & Joan Didion          & 39 & Roxane Gay (\checkmark)\\
 4 & Ben Lerner (\checkmark)            & 22 & Jonathan Franzen (\checkmark)    & 40 & Sally Rooney (\checkmark) \\
 5 & Charles Bukowski       & 23 & Junot Díaz (\checkmark)          & 41 & Salman Rushdie (\checkmark)\\
 6 & Cheryl Strayed (\checkmark)        & 24 & Kazuo Ishiguro (\checkmark)       & 42 & Shirley Jackson \\
 7 & Chimamanda Ngozi Adichie (\checkmark) &25 & Louise Erdrich (\checkmark)    & 43 & Sigrid Nunez (\checkmark) \\
 8 & Colson Whitehead  (\checkmark)     & 26 & Lydia Davis         (\checkmark) & 44 & Stephen King \\
 9 & Cormac McCarthy        & 27 & Margaret Atwood (\checkmark)     & 45 & Tony Tulathimutte (\checkmark)\\
10 & David Foster Wallace   & 28 & Marilynne Robinson (\checkmark)   & 46 & V. S. Naipaul \\
11 & Ernest Hemingway       & 29 & Maya Angelou         & 47 & Virginia Woolf \\
12 & Flannery O'Connor      & 30 & Milan Kundera        & 48 & William Faulkner \\
13 & Gabriel García Márquez & 31 & Min Jin Lee  (\checkmark)        & 49 & Yoko Ogawa (\checkmark)\\
14 & George Saunders (\checkmark)       & 32 & Nora Ephron          & 50 & Zadie Smith (\checkmark)\\
15 & Han Kang (\checkmark)              & 33 & Octavia Butler       &    & \\
16 & Haruki Murakami (\checkmark)       & 34 & Orhan Pamuk (\checkmark)         &    & \\
17 & Hunter S. Thompson     & 35 & Ottessa Moshfegh (\checkmark)    &    & \\
18 & Ian McEwan  (\checkmark)           & 36 & Percival Everett (\checkmark)     &    & \\
\bottomrule
\end{tabular}
\vspace{1ex}
\caption{\label{tab:authorlist}Author list for our pool of 50 authors. (\checkmark) denotes authors who were used in the fine-tuning experiment.}
\end{table*}

\begin{figure*}[!htbp]
    \centering
    \includegraphics[width=0.85\textwidth]{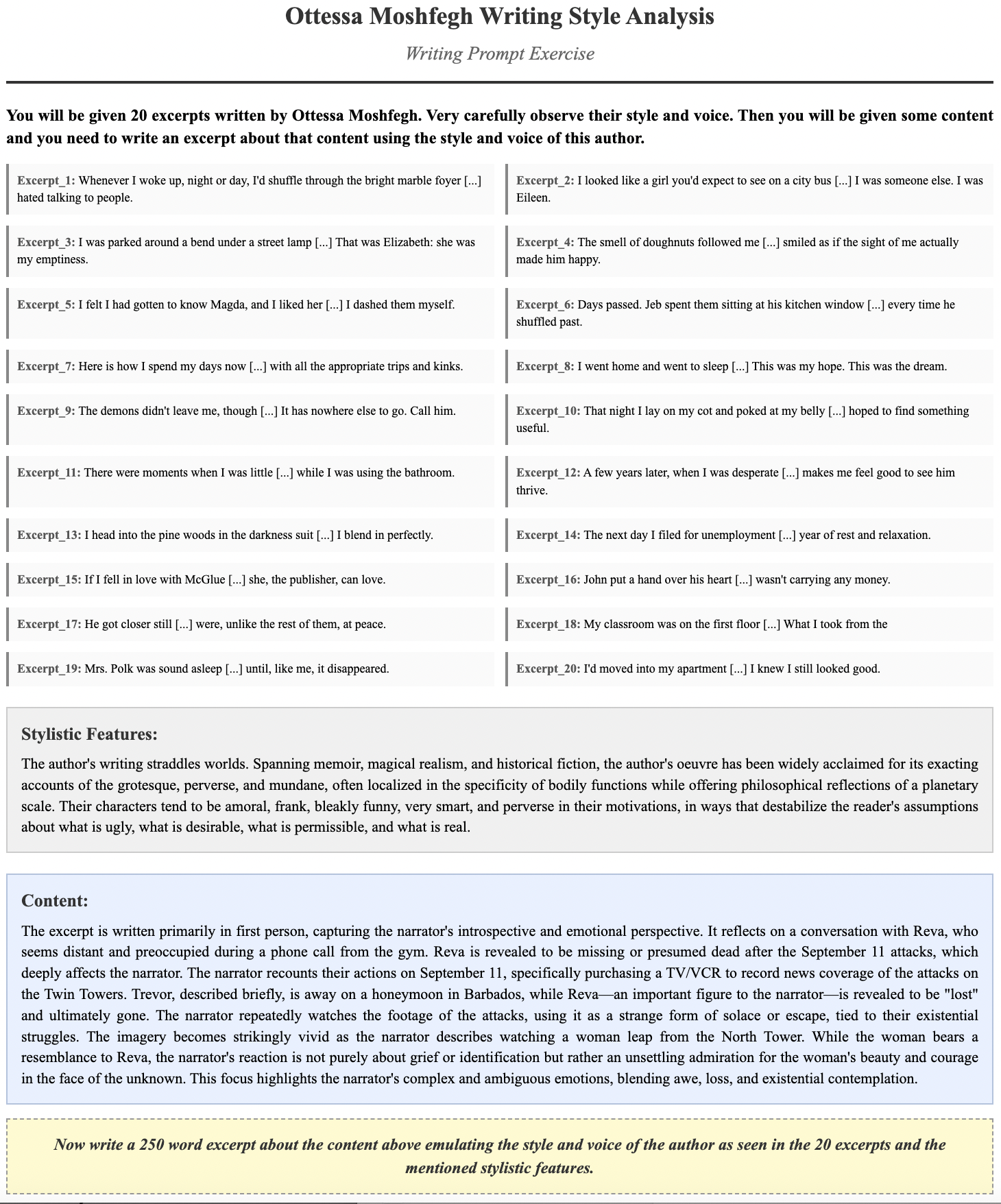}
    \caption{Writing Prompt to emulate Ottessa Moshfegh's style/voice}
    \Description{Writing prompt document titled "Ottessa Moshfegh Writing Style Analysis." The prompt presents 20 numbered excerpts from Moshfegh's works arranged in two columns, followed by a "Stylistic Features" section describing her writing as spanning memoir, magical realism, and historical fiction with amoral, frank, and bleakly funny characters. A "Content" section describes a specific scene involving the narrator, a character named Reva, and September 11 imagery. The prompt concludes with instructions highlighted in yellow: "Now write a 250 word excerpt about the content above emulating the style and voice of the author as seen in the 20 excerpts and the mentioned stylistic features."}
    \label{fig:writingprompt}
\end{figure*}

\subsubsection{Designing writing task for humans}
As mentioned earlier, to improve AI writing, practitioners increasingly prompt AI systems to perform style mimicry by emulating specific writers' choices ~\cite{chiang_why_2024}. While the effectiveness of such stylistic emulation remains contested, the more pressing question concerns whether style mimicry genuinely improves the quality of AI-generated text and whether judges perceive these improvements as meaningful. To address this question, in collaboration with 5 English Ph.D. students we selected 50 internationally acclaimed published authors with distinct literary style/voice (See Table \ref{tab:authorlist}). To ensure consistency in the writing task, we provided each writer with an author-specific writing prompt. Each prompt consisted of three components: (i) twenty sample excerpts spanning an author's complete body of work, (ii) textual descriptions of the author's distinctive style and voice, and (iii) detailed content specifications about the original author-written excerpt that participants were required to emulate. These writing prompts were developed in collaboration with the 5 English Literature Ph.D. students who analyzed each author's literary voice and created the verbalized style descriptions. The 20 excerpts from the authors were chosen manually after closer inspection for the in-context prompt. Figure \ref{fig:writingprompt} shows an example of a writing prompt. Writers could choose the author they wished to emulate and were compensated \$75 for each writing task. The excerpts in this task ranged from 200-450 words. Each author was assigned to exactly three writers to ensure balanced representation. In terms of length of the excerpts we had strict requirements around adhering to the length of the original author-written excerpts. For 50 authors we thereby obtained a total of 150 excerpts. Almost every writer wrote more than two excerpts for the purpose of the study. Refer to Figure \ref{fig:themes} in Appendix for the themes of the emulated excerpts. For the purpose of the study, MFA-trained expert writers had no time limitation to submit these excerpts and were encouraged to take as long as required to produce their best work.
\subsubsection{Designing the writing task for LLMs\label{sec:llmwriting}}
Like human writers we asked LLMs to perform the same writing task. In particular our experiments involved two approaches i) In-context Prompting ii) Fine-tuning. See Figure \ref{fig:study_findings} for more details.
\paragraph{\textbf{In-context Prompting:}} Our first experiment examined prompting-based style/voice emulation. One of the most popular ways to get any desired output from LLMs has been via in-context Prompting where the model is first shown some demonstration of the task before performing it. We converted the identical writing prompt provided to writers into long-context prompts for three widely used large language models: OpenAI's GPT-4o, Anthropic's Claude 3.5 Sonnet, and Google's Gemini 1.5 Pro. To make it consistent with human writers we generated one excerpt per author for each LLM. Therefore we obtain a total of 150 LLM generated excerpts (50 from each LLM).
\begin{figure*}[!htbp]
    \centering
    \includegraphics[width=0.85\textwidth]{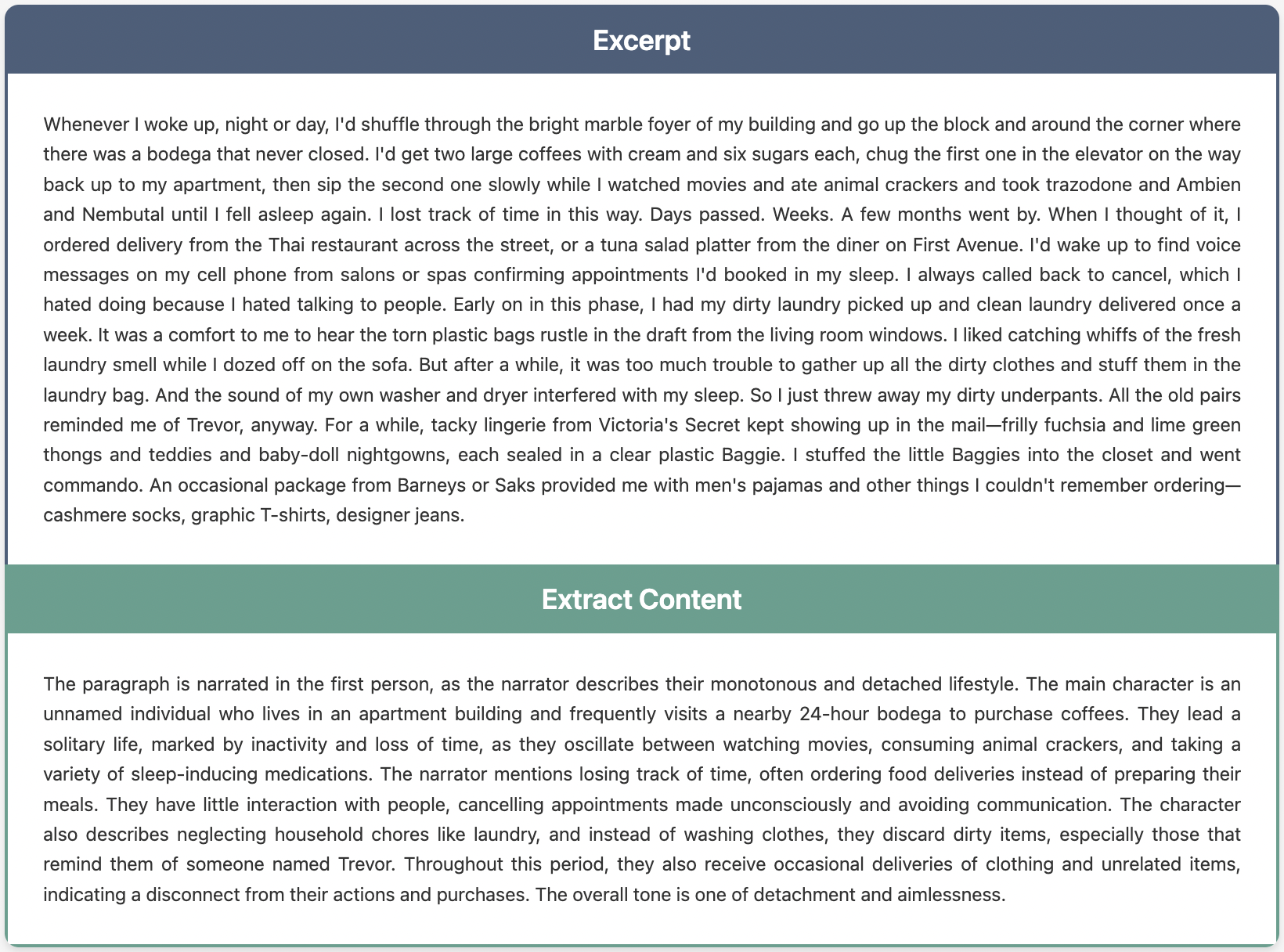}
    \caption{Segmented excerpt and extracted content from the excerpt}
    \label{fig:extractcontent}
    \Description{Two-panel figure showing an excerpt and its extracted content. The top panel, labeled "Excerpt," contains a paragraph from Ottessa Moshfegh's "My Year of Rest and Relaxation" describing the narrator's detached daily routine of visiting a bodega for coffee, taking sleep medications, ordering food delivery, and discarding belongings. The bottom panel, labeled "Extract Content," provides a content summary describing the first-person narration of a monotonous, solitary lifestyle marked by inactivity, avoidance of human interaction, and an overall tone of detachment and aimlessness.}
\end{figure*}

\paragraph{\textbf{Fine-tuning:}}We selected 30 living authors from our group of 50 for the purpose of fine-tuning (See Table \ref{tab:authorlist}). This choice was driven by our goal to focus on currently existing authors whose styles remain culturally relevant, while also taking into account the significant cost to buy ebooks written by these authors and financial cost  required to train models on each author's works individually. We bought digital ePub versions of these authors' books and transformed them into plain text files. 

\begin{figure*}[!htbp]
    \includegraphics[width=0.9\linewidth]{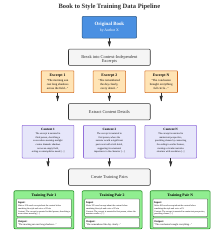}
    \caption{\label{fine-tuning}The pipeline used to fine-tune GPT-4o on an author's entire oeuvre.}
    \Description{Flowchart illustrating the fine-tuning pipeline for GPT-4o on an author's oeuvre. The process has four stages: (1) Original book by Author X is broken into context-independent excerpts, (2) Multiple excerpts (1 through N) are extracted, (3) Content details are extracted from each excerpt describing narrative perspective and scene elements, (4) Training pairs are created with inputs containing style-emulation prompts and content descriptions, paired with the original excerpt text as outputs.}
\end{figure*}

The only frontier model (among the 3 models used for In-context Prompting setting) that allows fine-tuning via API is ChatGPT (GPT-4o), so we used that. We segmented an entire book into context-independent excerpts. To begin with, we split the book text at existing double-newlines and rejoin them to maintain excerpt size limits (250-650 words). In rare instances where this naive approach produces excerpts exceeding 650 words, we employ GPT-4o for further segmentation.\footnote{\textit{Segment it into excerpts of minimum length 300-350 words such that each excerpt is grammatical from the start and doesn't feel abruptly cut off. There should be zero deletion and break into excerpts at grammatically natural places. Maintain the original word count. Avoid breaking into too many small excerpts. Start directly. Don't say Here's or Here is....}} Following segmentation of book into excerpts, we extract content details using GPT-4o with the prompt: \textit{Describe in detail what is happening in this excerpt. Mention the characters and whether the voice is in first or third person for majority of the excerpt. Maintain the order of sentences while describing.} Figure \ref{fig:extractcontent} displays a sample paragraph from My Year of Rest and Relaxation by Ottessa Moshfegh alongside its extracted content. After obtaining these content descriptions, we fine-tune GPT-4o\footnote{Version: gpt-4o-2024-08-06} through their fine-tuning API using the input-output pair: \textbf{Write a [[n]] word excerpt about the content below emulating the style and voice of [[authorname]] \textbackslash n \textbackslash n[[content]]: [[excerpt]]}. This method is commonly known as instruction back-translation \cite{li2023self}. We excluded the original author-written excerpt from the training set for fairness purposes. See Figure \ref{fine-tuning} for the full workflow/pipeline. During inference we ensured that none of the excerpt generated across both AI conditions regurgitated verbatim expressions from the original author-written excerpt. ROUGE-L scores \cite{lin-2004-rouge}\footnote{measures the longest common subsequence between the generated text and reference text } (ranging from $\sim$0.16 to $\sim$0.23) indicate relatively very low overlap between the AI-generated text and the original author-written excerpt.

\begin{figure*}[t]
    \centering
    \includegraphics[width=0.85\textwidth]{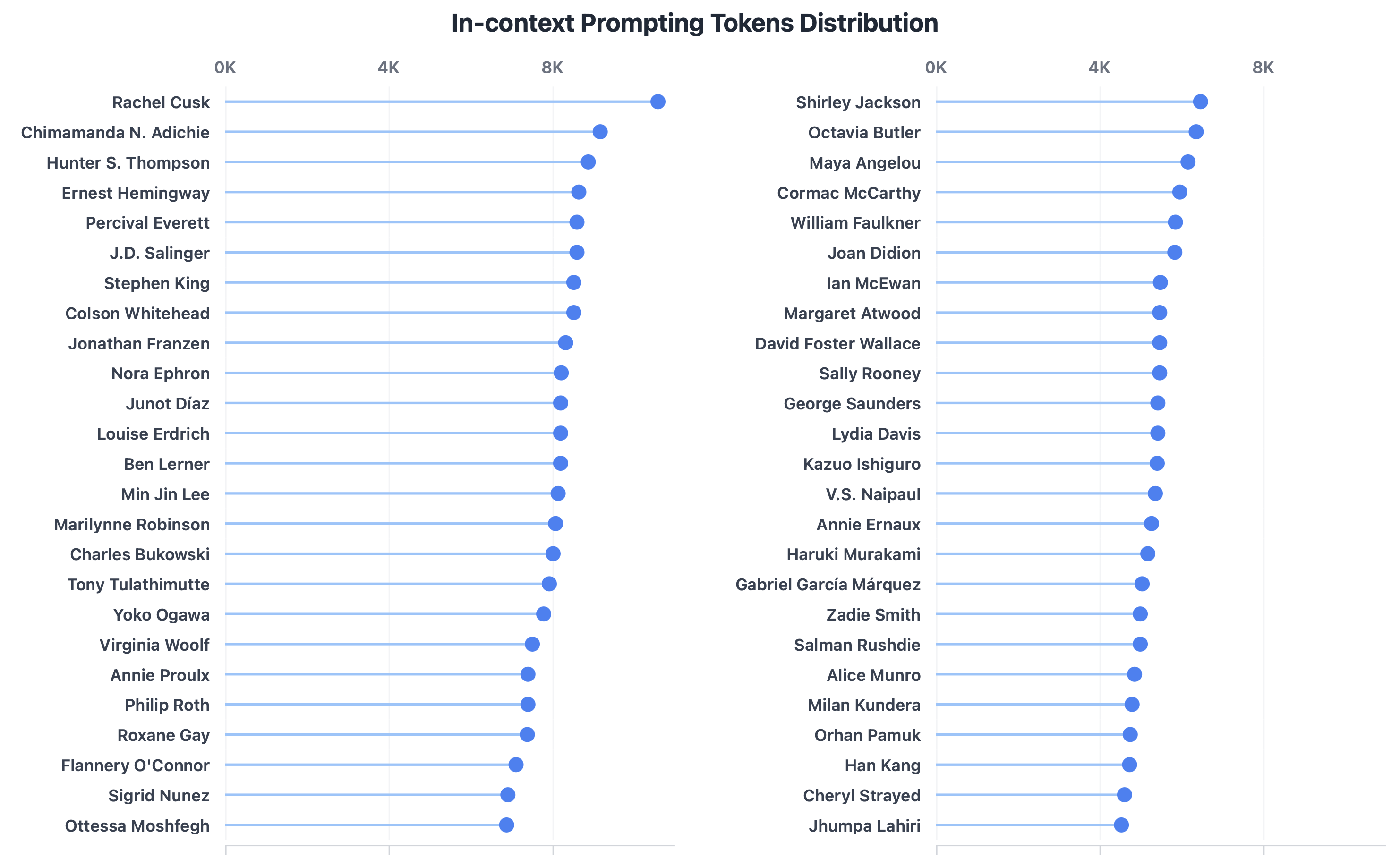}
    \caption{Number of tokens used for in-context prompting.}
    \Description{Lollipop chart displaying the number of tokens used for in-context prompting across 50 authors, arranged in two columns of 25 authors each. Authors are sorted by token count in descending order, with horizontal lines extending from author names to circular markers indicating token values. The scale ranges from 0 to approximately 11K tokens, with values varying from around 4.5K to 10.6K tokens per author.}
    \label{fig:size-incontext}
\end{figure*}

\begin{figure}[htbp]
    \centering
    \includegraphics[width=0.4\textwidth]{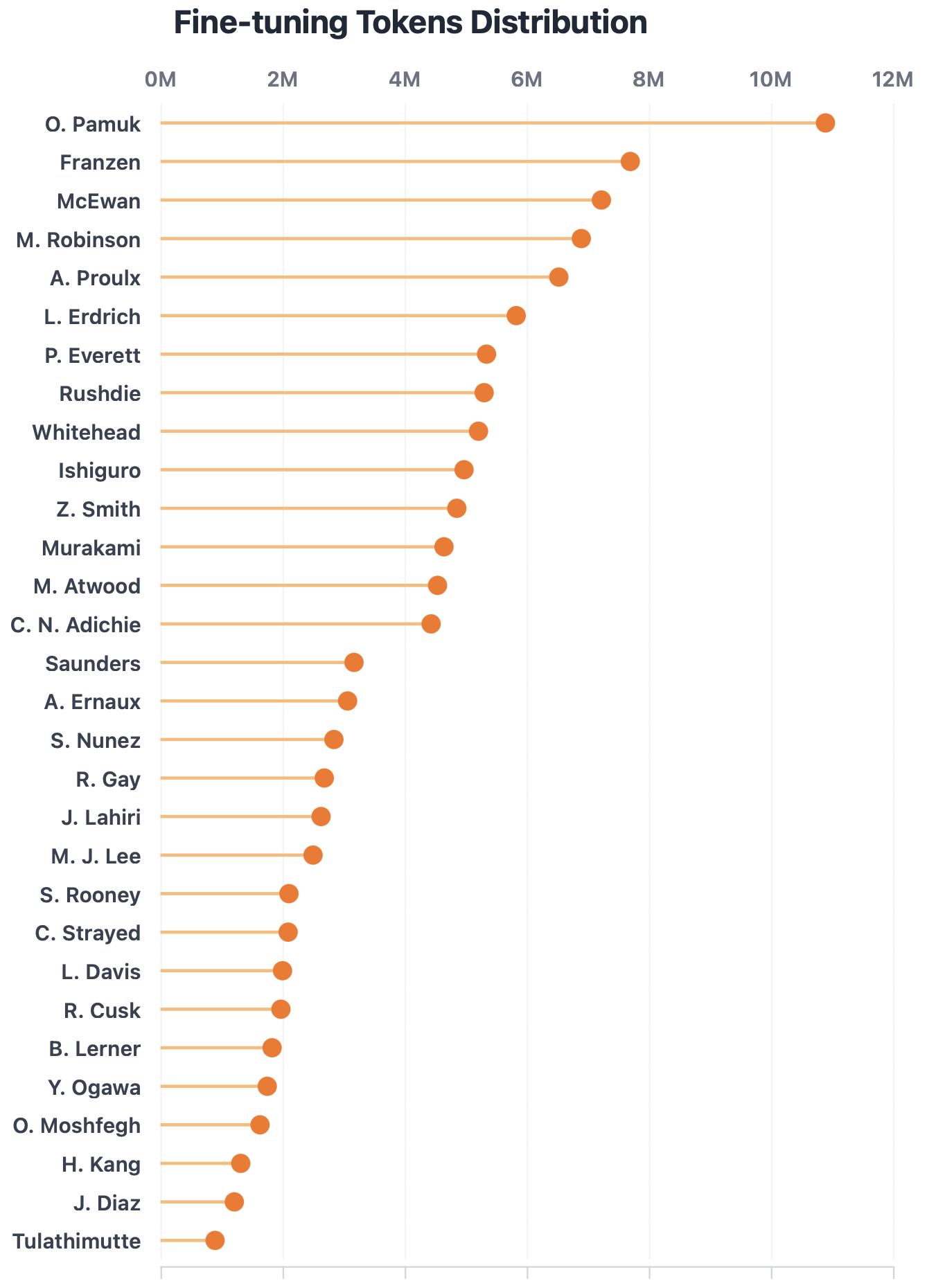}
    \caption{Number of training tokens for Fine-tuning. Fine-tuning requires 583 times more tokens on average compared to prompting (cf. Figure~\ref{fig:size-incontext})}
    \Description{Lollipop chart displaying the number of training tokens used for fine-tuning across 30 authors in a single column. Authors are sorted by token count in descending order, with horizontal lines extending from author names to circular markers indicating token values. The scale ranges from 0 to 12 million tokens, with values varying from approximately 0.9M to 10.9M tokens per author. Fine-tuning requires 583 times more tokens on average compared to in-context prompting.}
    \label{fig:size-finetune}
\end{figure}

There is a significant difference in terms of how data intensive the two different LLM conditions are. Figures \ref{fig:size-incontext},~\ref{fig:size-finetune} show that fine-tuning requires 583 times more number of tokens on average compared to In-context Prompting. For the in-context prompting setup, we therefore end up with 150 $<Human, AI>$ pairs and for fine-tuning setup we end up with 90 $<Human, AI>$ pairs.

\vspace{-1ex}
\subsubsection{Choice and Nature of Writing Task}
Coming up with ideas or content and putting them into words fundamentally requires style and voice because the same content becomes radically different literature depending on how it is rendered. Edward P. J. Corbett's classic article The Theory and Practice of Imitation in Classical Rhetoric \cite{corbett1971theory} documents how imitation was central to learning to write and speak effectively, and later composition scholars have argued for a revival of sentence- and style-imitation exercises in contemporary teaching. Eugène-Melchior de Vogüé, discussing Russian realist writers, said: \textit{We all came out from under Gogol's Overcoat.} Developing one's own distinct literary style and voice often takes years, and many young writers subconsciously borrow from other authors' styles. In the Paris Review, American author Benjamin Nugent acknowledged that he spent much of his twenties trying to imitate George Saunders \cite{nugent2020imitateSaunders}. This has also been acknowledged by George Saunders himself, who teaches at Syracuse. In his words:  \textit{Kids come [into the Creative Writing program] and imitate other writers} \cite{vanderark2018saundersLearningWrite}. This made MFA-trained writers an ideal choice for our task; while developing their own voices, they are heavily influenced by other writers. We also gave our MFA-trained expert writers the choice of which author they wanted to emulate. This reduced any possible friction and made the writing task effortless and enjoyable.

While our task doesn't require participants to write an original story in their own voice, it should still be noted that our writing task requires originality under explicit content constraints. Although both AI and human writers received prompts with detailed content specifications, composing literary fiction and creative nonfiction still requires agency—especially in deliberate choices of words, syntax, voice, and narrative framing—to produce novel, coherent prose. Consider the Marilynne Robinson excerpt in Table \ref{emulation}: the prose achieves its effect not through plot mechanics but through the accretion of syntactic choices such as the cascading dependent clauses, the modulation between concrete image and abstract reflection as well as the deliberate withholding of punctuated closure. These are not features one can specify in a prompt; they emerge from compositional judgment exercised sentence by sentence. Because literary fiction and creative nonfiction are inherently non-formulaic, the writing task extends beyond style imitation to creative composition, making it different from mere parody or pastiche.

Our study examines shorter (up to 450-word) excerpts and because of this, broader creative elements such as plot, character development, or structure might not be accurately reflected. We however focus on this length for two reasons. First, an excerpt of this length isolates the voice, style, and quality that writers must sustain across any longer work. Second, in its current form, an LLM cannot write long-form narratives that automatically balance coherence, quality, thematic consistency, and plot development across thousands of words. But it is merely a matter of time before LLMs reach that capacity. Even now, with human steering and iterative prompting, it is feasible to produce long-form fiction and nonfiction using shorter model-generated excerpts. This is already happening in real life, with startups such as Sudowrite or genre fiction publisher Inkitt \cite{varaairomancefactory2025} specifically focusing on helping consumers write full books using AI. As shown in prior work \cite{jonesaiknockoffs2025}, in self-publishing marketplaces such as Kindle, these AI-generated books have already gained popularity \cite{knibbsscammyaibooks2024}.

\subsection{Evaluation Task}
\subsubsection{Recruitment of Judges} For evaluation purposes, we rely on both experts as well as the average lay human as judges. We recruit the same 28 writers who act as expert judges. All the writers recruited in our study have teaching experience in the US, as it's mandatory for them as a part of their MFA curriculum to act as instructors for undergraduate writing classes. Expert writers however aren't always the representative consumer base. For this purpose we rely on 131 lay judges recruited from Prolific. Given the challenging nature of the evaluation task, we restricted ourselves to English-speaking countries (USA and UK). We also required participants to be born in these countries and have a 100\% acceptance rate, be college-educated. It should be noted that expert judges (who served as participants in the writing task) never evaluated their own work.
\subsubsection{Evaluation Setup}

As shown in Figure \ref{fig:study_findings} Phase 2, evaluators are shown a pair of  $<Human_{written}$,$AI_{generated}>$ excerpts centered around the exact same content for Writing Quality evaluation. Because style is judged with respect to a reference for Stylistic Fidelity evaluation, evaluators are shown a triple of $<Original, Human_{written}$, $AI_{generated}>$. We never notified our participants at any point during the study whether any given text was human-written or AI-generated due to the psychological bias associated with AI-writing~\cite{li-etal-2024-disclosure,sarkar2025ai,ling2025underreporting}. They are then required to submit their preference about which excerpt has better writing quality and which excerpt is closer to the original in terms of stylistic fidelity. Prior work \cite{mcdonnell2016relevant} has shown that requiring annotators to supply a rationale with each judgment increases transparency into how the decision was made and supports quality control/adjudication. Motivated by this, we asked both expert and lay judges to provide 2-5 sentence rationales grounded in snippets from these excerpts, explaining their preference (See \ref{fig:prefeval} in Appendix for more detail). These preferences along with rationales helped us understand both quantitatively and qualitatively the extent to which experts and lay judges agree or disagree on what qualifies as ``good writing" or which excerpt is closer to the style of the original author and, more specifically, what lines of reasoning each group uses for their respective preferences. We paid participants \$75 to evaluate a batch of 10 pairs of excerpts for Writing Quality and \$100 for a batch of 10 for Stylistic Fidelity. Three distinct experts and five distinct lay judges evaluated each $<Human\_written, AI\_generated>$ pair or $<Original, Human_{written}$, $AI_{generated}>$ triple with the final decision determined through majority voting. Writers who acted as expert judges didn’t assess their own work. Figures \ref{fig:evaluation_screen} and \ref{fig:evaluation_screen1} (in the Appendix) show the evaluation interfaces for each task. A curious reader might wonder why we didn't compare the AI-generated excerpt to an excerpt written by the original human author. We couldn't do this because the original human author's work could be found on the internet via Google Search, or if the participants had read the original excerpt before, they would be biased to choose the original version.

\subsection{Debrief}

\begin{table*}[!htbp]
\centering
\small
\begin{tabular}{|lll|}
\hline
\multicolumn{3}{|l|}{\textbf{Exploring prior relationship with technology}}                                                                                                                                                                                                          \\ \hline
\multicolumn{3}{|l|}{\begin{tabular}[c]{@{}l@{}}\textit{Before we discuss the evaluation results, how have technological tools shaped your writing and what are your thoughts}\\ \textit{on AI and creative writing?}\end{tabular}}                                                           \\ \hline\hline
\multicolumn{3}{|l|}{\textbf{Revelation}}                                                                                                                                                                                                                                            \\ \hline
\multicolumn{3}{|l|}{\textit{What is your first thought upon realizing your preferences?}}                                                                                                                                                                                          \\ \hline\hline
\multicolumn{3}{|l|}{\textbf{Sense-making questions}}                                                                                                                                                                                                                                \\ \hline
\multicolumn{3}{|l|}{\begin{tabular}[c]{@{}l@{}}\textit{"I'd like to read back what you wrote about why you preferred this piece. Now knowing it was AI-generated, how do}\\ \textit{you feel about these reasons?"}\end{tabular}}                                                           \\ \hline
\multicolumn{3}{|l|}{\begin{tabular}[c]{@{}l@{}}\textit{"You mentioned the writing had {[}specific quality they praised - e.g., 'emotional resonance,' 'vivid imagery,' 'authentic}\\  \textit{voice'{]}. What does it mean to you that you attributed these qualities to AI?"}\end{tabular}} \\ \hline\hline
\multicolumn{3}{|l|}{\textbf{Identity impact questions}}                                                                                                                                                                                                                             \\ \hline
\multicolumn{3}{|l|}{\textit{How did this discovery relate to your sense of yourself as someone who recognizes good writing?}}                                                                                                                                                     \\ \hline
\multicolumn{3}{|l|}{\textit{What did it mean to you that an AI could create something you found compelling?}}                                                                                                                                                                      \\ \hline
\multicolumn{3}{|l|}{\textit{Has this affected your confidence in your ability to evaluate writing?}}                                                                                                                                                                                \\ \hline\hline
\multicolumn{3}{|l|}{\bf Practice evolution questions}                                                                                                                                                                                                                          \\ \hline
\multicolumn{3}{|l|}{\textit{How do you now think about AI's role in creative work?}}                                                                                                                                                                                               \\ \hline\hline
\multicolumn{3}{|l|}{\textbf{Future of Writing}}                                                                                                                                                                                                                                 \\ \hline
\multicolumn{3}{|l|}{\textit{What does this mean for the value of MFA training?}}                                                                                                                                                                                                   \\ \hline
\multicolumn{3}{|l|}{\textit{How might the literary establishment need to evolve?}}                                                                                                                                                                                               \\ \hline
\end{tabular}
\vspace{1ex}
\caption{\label{questionnaire}Pre-determined questionnaire for Debrief interviews}
\end{table*}

A few weeks after the entire evaluation task was concluded, we asked our writers to participate in a debrief (See Figure \ref{fig:study_findings} Phase 3). 21 out of 28 participants responded to participate in the debrief. The first author of the paper conducted hour-long virtual interviews over Google Meet with our writers. The video was transcribed automatically and then errors were manually fixed by the authors. The transcripts were read by all authors of the paper. These debrief interviews followed a \textbf{retrospective semi-structured} format \cite{patton2014qualitative}, that allowed our participants to reflect on their experiences with full knowledge of the study's design while exploring their reactions to receiving critiques and critiquing both human and AI-generated work. This approach combined elements of \textbf{member checking} (validating interpretations with participants) \cite{lincoln1985naturalistic,birt2016member} with \textbf{reflexive interviewing} \cite{finlay2002negotiating,denzin2001reflexive}, where participants co-analyze in making sense of the collective patterns and implications emerging from the study data.

Prior to interviews our team discussed together and decided on a list of pertinent questions as shown in Table \ref{questionnaire}. We also shared a spreadsheet with 15 $<${\it Excerpt1, Excerpt2, Preference, Rationale, Agree/Disagree/Partially Agree, Additional Comments}$>$ tuples to every writer a few days prior to the interview. 10 of the first 15 excerpt pairs were written by them where an evaluator had chosen an AI to be of better quality over their own writing. The last 5 excerpt pairs were ones where they had chosen an AI excerpt to be of better writing quality over one written by another writer. These excerpts were sampled randomly from both In-context and fine-tuned setup.

\begin{table*}[!ht]
\centering
\small
\renewcommand{\arraystretch}{1}
\begin{tabular}{|ll|}
\hline
\multicolumn{2}{|l|}{\begin{tabular}[c]{@{}l@{}}\textit{Sometimes I wonder about the difference between craving something and fulfillment. Sometimes I want something so much}\\ \textit{I feel like I’m going to break open. Let’s say it’s a fresh berry, let’s say it’s like I’ve been really wanting a fresh berry, plucked}\\ \textit{from a bush in the orchard, even before that berry’s in my hands, it’s already got a taste in my mind, I’ve been building it up} \\ \textit{moment after moment until I have it in my brain like I’ve already got it on my tongue, so when Sylvie tells me she picked} \\ \textit{something for me and plops it in my mouth, it’s even sweeter because the sweetness confirms the foundation I’ve already built}\\\textit{ for it. I don’t know how to put it. Sylvie disappeared for a while, like a mouse, but then she came back. Sometimes she still} \\ \textit{goes out though, vanishes into the woods without saying anything. I think she’s out there right now, watching me ruminate. I}\\ \textit{don’t know what kind of solitude we can both get when we both are in each other’s minds. Could it be a way of being} \\ \textit{together? A very strange way? It’s beautiful out in the woods. The trees are older than I can fathom, really, old like stories} \\ \textit{can be old. I’d like to put a status out there, almost like somebody would always be watching, a statue of a woman, to stand in} \\ \textit{between all those old trees, a woman in a crown of new flowers, looking only ever ahead, forward, never letting any kind of} \\ \textit{barrenness touch her, like she was the opposite of Lot’s wife and all there was to her was the kind of forgiving vibrancy that} \\ \textit{children wonder at, the vibrancy that comes from a life well-lived. Even as a statue, I think they’d come to her, every child} \\ \textit{that’s ever felt lost-wild and orphan, and I think when they’d reach out to touch her stone hands, they’d find that that statue’s} \\ \textit{stone was warm for them. Sometimes I wish Sylvie were warmer. There’s warmth all around in all the beautiful trees and} \\ \textit{even the grasses. The problem is that trees don’t feel warm when you touch their bark unless it’s from the sun.}\end{tabular}} \\ \hline\hline
\multicolumn{2}{|l|}{\begin{tabular}[c]{@{}l@{}}\\ \textit{Craving and its indulgence are really one and the same thing. There might be no fruits, and still, if one wished for them,} \\ \textit{yearned for them, one would taste them all. The very sight of a berry in the wild is an admonishment not to pick it, for when} \\ \textit{it is stored in the vision and played upon by desire, the taste it bears is gifted to it as much by memory as by its savor. The} \\ \textit{flavor grows until it is too exquisite for the tongue, unencumbered by flesh. So it is when one wishes for a friend, or a}\\ \textit{woman, or a child. For Sylvie disappeared and made no sound, and I fancied when I had closed my eyes and rocked awhile }\\ \textit{that she was not gone. It was easier than I could have imagined to pretend I was not alone, to insist that she merely watched me} \\ \textit{from the woods. The wind was dying, the sun was sinking, the side of the house was cooling, the evening moths were glancing} \\ \textit{and settling, and her absence was a thing so full of longing it became a presence of its own. If I could row the lake at dawn and} \\ \textit{gather all the proud blossoms from the woods for one year, I would throw some across the graves of unsanctified and} \\ \textit{dreamless children and some into the holes of fox dens. And with the rest I would sculpt from snow a statue of a woman. I} \\ \textit{would stand her in the clearing in the woods, and she would be adorned with rare flowers with leaves like hands, with hard-stem} \\ \textit{thistles, and the birds would rest in her. Lot's wife would have none of this: flowers in profusion and children. Her breast caved} \\ \textit{into a hollow of regret. Not this one, not my mother. All the wild children, who knew too well that the world does not give itself} \\ \textit{generously to children, would burst from the trees and behold her. And every child, the most orphaned, the most bereft, would} \\ \textit{forgive her for her frozen hands and would stand at her white hem in awe that she should seem to shower them with blossoms,} \\ \textit{eternally offering what she could never give, inexplicably so stern.}\end{tabular}}                                       \\ \hline
\end{tabular}
\vspace{2ex}
\caption{\label{emulation}Marilynne Robinson Emulation: Human-Written excerpt on Top. Bottom: Fine-tuned AI excerpt.}
\end{table*}

\subsection{Institutional Review Board Approval, Participant privacy and Advocacy Statement}
Our study was approved by University of Michigan IRB (HUM00264127). Informed consent was collected from all participants. No data provided by writers in the study is shared publicly. All evaluation data and interview files are saved on secure servers. While we train models on legally obtained copyright-protected books for the purpose of the study, as authors of the paper we do not advocate for this. As a matter of fact, our results highlight the risk of fine-tuning models on high-quality data that often comes from copyrighted books. We purchased ebooks for 30 authors from \url{https://www.ebooks.com/} and processed and saved data on secure servers. All fine-tuned GPT‑4o models belong to first author's personal account and cannot be distributed or shared publicly. We do not plan to release fine-tuned models but we will share steps for reproducibility. Additionally we obtained legal advice and approval from top US copyright scholars who agreed that any scanning of lawfully acquired books for academic research purposes would be fair use under the circumstances.

\section{Findings}
\subsection{How do human writers compare to AI?}

We analyzed the final preference data based on majority voting from our study comparing human (MFA) written excerpts versus AI-generated ones across two conditions (In-context and Fine-tuned) and two evaluator groups (Expert and Lay). The data consisted of binary choices a) In-context condition: N=150 $<${\it Human\_written, AI\_generated}$>$ and $<${\it Original, Human\_written, AI\_generated$>$} evaluations per group (Expert and Lay) for Writing Quality and Stylistic Fidelity b) Fine-tuned condition: N=90 $<${\it Human\_written, AI\_generated}$>$ evaluations per group (Expert and Lay). See Table \ref{emulation} for a sample $<${\it Human\_written, AI\_generated}$>$ pair.

\begin{figure*}[htbp]
    \centering
    \includegraphics[width=\textwidth]{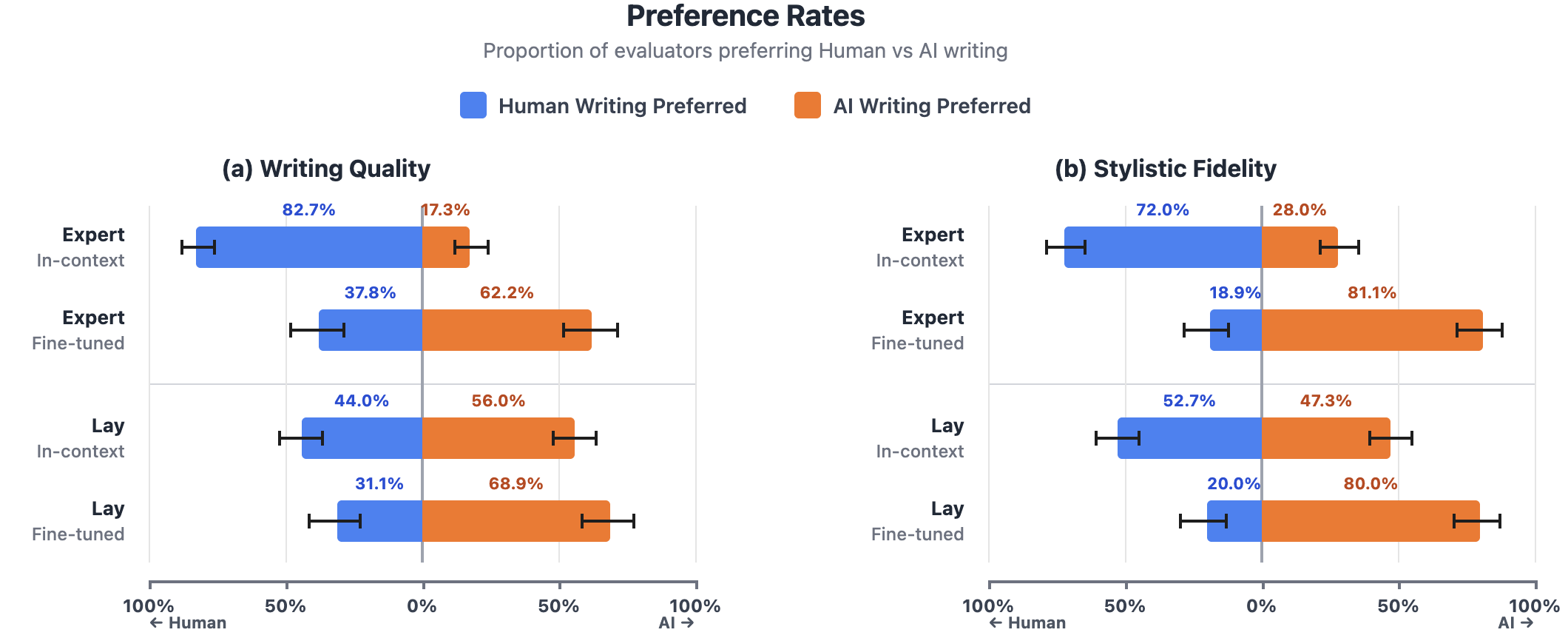}
    \caption{\textbf{Preference Rates for Writing Quality and Stylistic Fidelity.} Diverging bar charts showing the proportion of evaluators preferring human-written versus AI-generated text for (a) Writing Quality and (b) Stylistic Fidelity. Bars extend left for human preference (blue) and right for AI preference (orange); the center line represents equal preference. \textbf{(a) Writing Quality:} In the In-context condition, experts strongly preferred human writing (82.7\%) while lay evaluators showed no clear preference (44.0\%). In the Fine-tuned condition, both experts (62.2\%) and lay evaluators (68.9\%) preferred AI-generated writing. \textbf{(b) Stylistic Fidelity:} In the In-context condition, experts preferred human writing (72.0\%) more than lay evaluators (52.7\%). In the Fine-tuned condition, both experts (81.1\%) and lay evaluators (80.0\%) strongly preferred AI-generated writing. Statistical comparisons are reported in Section 4.1.}    
    \label{fig:preference-rates}
    \Description{Side-by-side diverging bar charts showing preference rates for (a) writing quality and (b) stylistic fidelity. Each chart displays four horizontal bars for expert and lay evaluators under In-context and Fine-tuned conditions. Blue bars extend left for human preference; orange bars extend right for AI preference. Writing Quality (left panel): Expert In-context shows 82.7\% human preference (long blue bar); Lay In-context shows 44.0\% (short blue bar near center); Expert Fine-tuned shows 37.8\% (orange bar right); Lay Fine-tuned shows 31.1\% (orange bar right). Stylistic Fidelity (right panel): Expert In-context shows 72.0\% human preference (blue bar left); Lay In-context shows 52.7\% (shorter blue bar); Expert Fine-tuned shows 18.9\% (long orange bar right); Lay Fine-tuned shows 20.0\% (long orange bar right). Error bars show 95\% confidence intervals. Both charts demonstrate dramatic shifts from human to AI preference when moving from In-context to Fine-tuned conditions, with expert evaluators showing larger shifts than lay evaluators, and stylistic fidelity showing larger shifts than writing quality.}
\end{figure*}

We performed chi-square tests of independence to examine \textbf{within-condition comparisons}, i.e., whether expert and lay evaluators differed in their preferences within each condition, and \textbf{between-condition comparisons}, i.e., whether preferences changed between In-context and Fine-tuned conditions for each evaluator group. We calculated Cramér's V as a measure of effect size for each comparison, using the formula $V = \sqrt{\frac{\chi^2}{n}}$ where $n$ is the total number of observations in the contingency table.\footnote{Effect sizes were interpreted using Cohen's conventions: small ($V = 0.1$), medium ($V = 0.3$), and large ($V = 0.5$).}

Expertise matters in the In-context condition: expert judges strongly preferred human writing (82.7\% for Writing Quality, 72.0\% for Stylistic Fidelity), while lay judges showed weaker or no clear preference (44.0\% and 52.7\%, respectively). This within-condition difference was statistically significant for both metrics ($p < 0.001$). Fine-tuning dramatically shifts these patterns. After fine-tuning, both groups preferred AI-generated writing, with no significant difference between evaluator types ($p > 0.05$). The expert-lay gap shrinks from 38.7\% to 6.7\% for Writing Quality and from 19.3\% to $-1.1$\% for Stylistic Fidelity, suggesting that author-specific fine-tuning makes AI writing indistinguishable even to trained writers (Figure \ref{fig:preference-rates}). The effect is stronger for experts, who show larger preference shifts toward AI ($-44.9$\% for Writing Quality; $-53.1$\% for Stylistic Fidelity) than lay evaluators ($-12.9$\% and $-32.7$\%, respectively), possibly because fine-tuning eliminates the stylistic tics—clichés, awkward phrasing, overly ornamental language—that experts used to distinguish human from AI-generated text. Inter-annotator agreement also differed: experts reached moderate-to-substantial agreement ($\kappa = 0.41$--$0.67$ for Writing Quality; $\kappa = 0.54$--$0.58$ for Stylistic Fidelity), whereas lay judges remained inconsistent ($\kappa = 0.07$--$0.22$). These results highlight the importance of recruiting experts for complex, subjective tasks such as writing quality assessment. Refer to Section \ref{finegrained-humanai} in Appendix for granular model-level data.

\subsection{What line of reasoning do experts and lay judges use to justify what qualifies as good writing?}

Lay judges disproportionately reward writing based on whether it ``flows well", is ``more organized,” or whether ``easier to follow.” They frequently foreground clarity, conciseness, straightforwardness as the reason for their preference. Additionally they mention emotional impact in general terms (such as ``moving," ``powerful”, ``beautiful") more than the mechanisms that generate it (See Table \ref{rationales}). As a matter of fact for a David Foster Wallace emulation one lay judge wrote \textit{``Why are the sentences so unbearably long in Excerpt1? Is the author trying to force the reader to stress and strain over each word, compounding one after another to convey a strained meaning?"}. While understandable, David Foster Wallace is known for his maximalist and often very long sentences.
\begin{table*}[!htbp]
\centering
\renewcommand{\arraystretch}{1.15}
\small
\begin{tabular}{l|l}
Exp & \begin{tabular}[c]{@{}l@{}}The second excerpt has a much more idiosyncratic narrative voice, which allows some of Kant's own voice to seep through: \\ \textit{\color{blue}"he managed to send a taut, defensive email to a BCC list of family, friends, and coworkers, coming up with some bullshit} \\ \textit{\color{blue}Aristotelian lie about fairness and specifically that it's totally unfair to ever be in the closet at all."} This close third-person \\ voice allows the reader to better understand Kant as a character: his self-deprecation, humor, and internet milieu. The\\ internet vernacular in the second excerpt, also, helps situate Kant as an  internet-addicted millennial, an authenticity that's \\lacking in the first excerpt: \textit{\color{blue}"crashed out depressed on a coach"}, \textit{\color{blue}"turbocharged Like button"}, \textit{\color{blue}"silently clocked him."}\end{tabular}                                   \\ \hline
Lay    & \begin{tabular}[c]{@{}l@{}}I found the second excerpt hard to grasp at first, had to reread. The first is clear what is being communicated and less \\ conversational. The conversational nature of the second is hard to follow at times. the 1st uses simple language that is \\ also effective, fluent and coherent, unlike the second.\end{tabular}                                                                                                                      \\ \hline
Exp & \begin{tabular}[c]{@{}l@{}}Though about equal in coherence (each excerpt gets its message across quite well), Excerpt2 wins out in fluency and \\ effectiveness. We can see this in lines like, \textit{\color{blue}"The wind was dying, the sun was sinking, the side of the house was cooling} \\ \textit{\color{blue}, the evening moths were glancing and settling, and her absence was a thing so full of longing it became a presence} \\ \textit{\color{blue}of it's own"} the anaphora of which provides a sort of image-after-image punchiness that works quite well. Excerpt 1, \\ I feel, has a harder time finding its rhythmic footing, especially in the opening line and it's leaping associations that follow.\end{tabular}                                                                                     \\ \hline
Lay    & \begin{tabular}[c]{@{}l@{}}I feel that Excerpt 2 is coherent and very poetic \textit{\color{blue}"The very sight of a berry in the wild is an admonishment not to pick it,} \\ \textit{\color{blue}for when it is stored in the vision and played upon by desire, the taste it bears is gifted to it as much by memory as by its} \\ \textit{\color{blue}savor."} is a beautiful image in my opinion, and feels. Excerpt 1, whilst not bad, is weaker.\end{tabular}                                                                     \\ \hline
Exp & \begin{tabular}[c]{@{}l@{}}While I prefer some of the syntax of Excerpt1, as a whole I believe Excerpt2 to be more sophisticated and interesting. \\ In particular, I appreciate the way Excerpt2 unpacks the idea of collectivity and its relationship to artwork, that \textit{\color{blue}" one}\\\textit{\color{blue} wanted to be an artist to make bad forms of collectivity appear as the promise of an almost unimaginable good: not }\\\textit{\color{blue} reified abstraction, poorly disguised negation of the one in the many, but the emerging collective that would do justice} \\ \textit{\color{blue}to the wrongs absorbed by the bodies below the skyline."} Artwork, then, is like a mirror to a gnostic world, showing \\ something collective but true. By comparison, Excerpt1's depiction of this phenomenon seems rather simplistic. The \\ two sensations are not distinguished from each other, nor the presence of Noor from the whole: \textit{\color{blue}"All the bundled} \\ \textit{\color{blue}debt of midtown was merely a muse to a Greek woman’s pen; pigments intermingled in the city water supply with} \\ \textit{\color{blue}the traces of antidepressants; the gridlock on the Brooklyn Bridge recalled the stylings of Ibn Tulun."}\end{tabular} \\ \hline
Lay    & \begin{tabular}[c]{@{}l@{}}Excerpt 2 is very abstract \textit{\color{blue}"What I saw, I thought, was a material form of the second person plural, the true subject of art"} \\ which suits the subject matter and is effective in describing complex thoughts and ideas, but the flow of ideas is difficult \\ to follow. Excerpt 1 is easier to follow, but also more effective as setting the scene\end{tabular}  
\\
\end{tabular}
\vspace{1ex}
\caption{\label{rationales}Rationales written by Expert and Lay judges for same excerpt pairs}
\end{table*}

\begin{table*}[!htbp]
\centering
\renewcommand{\arraystretch}{1.05}
\small
\begin{tabular}{|l|l|}
\hline
\multirow{2}{*}{R1} & \begin{tabular}[c]{@{}l@{}}The \underline{\color{red}second excerpt uses overtly expository language} that nearly condescends to the reader, over-explaining the relationships \\ between the characters/their surroundings, though the characters' actions/dialogue speak for themselves: "It was clear that \\ both of us felt uncomfortable with the opulence of the space, but neither of us knew how to say so ... For a moment I imagined\\ myself as the girl, tan and comfortable among the trappings of wealth, though even then I knew that the hallmarks of the\\working class were written all over my face and my behavior." This "direct" overexplanation feels condescending to the reader,\\ and reduces the complexity of the characters' interactions to one-note themes, especially since this class difference is repeated\\ multiple times throughout the excerpt. The first excerpt, however, nicely leans on physical descriptions and the contrast between\\ the girl/the narrator to elucidate upon this idea: "The girl talked confidently to her father and laughingly spooned up her \\yogurt. \underline{\color{ForestGreen}In those} \underline{\color{ForestGreen}days, I didn't know what yogurt was."}\end{tabular}                                                                                                                   \\ \cline{2-2} 
                    & \begin{tabular}[c]{@{}l@{}}Excerpt 1 is better in terms of writing quality than Excerpt 2. It is well-balanced in terms of clarity and doubt on part of the \\ narrator, and how it provides information to the reader. Instead of \underline{\color{red}Excerpt 2's very expository sentence}, "My father and I \\ dined at an upscale restaurant...", Excerpt 1 keeps us guessing by keeping details obscured: "The restaurant was an annex...\\ I had a pale face...My father and I..." This involves the reader more actively in building the scene. Secondly, Excerpt 1 is more \\ exciting to read because of the narrator's admission of what they don't understand or know, which adds tension, and them trying \\ to guess what other characters are feeling. In addition, Excerpt 1 is more strategic in the sense that it introduces retrospection \\ in a more effective way by stating, through a separate sentence, \underline{\color{ForestGreen}"I didn't know what yogurt was,"} which adds a tinge of doubt \\ and sense of not knowing to the preceding sentence and creates the effect of knowledge arriving later. In Excerpt 2, on the other hand, \\ the exposition does not emulate the content so effectively because the narrator adds to the same sentence, the clause "I \\ later learned" after a comma.\end{tabular} \\ \hline
\multirow{2}{*}{R2} & \begin{tabular}[c]{@{}l@{}}Excerpt 1 is superior to Excerpt 2. Excerpt 1 has \underline{\color{ForestGreen}wonderfully specific and poetic imagery}, like "My mother's arms lay on top\\ of the blanket, waxy and unrecognizable to me now" and \underline{\color{ForestGreen}"Bluer than the enamel-blue tulip-shaped pot that Bruce would get} \\ \underline{\color{ForestGreen}in the mail a few days later."} Excerpt 2, on the other hand, is more general, filled with platitudes, and less interesting because of \\ it. Some example lines "It is one thing to know something and another thing to live it" and "A heroism shining in the blue of \\ her eyes, a deep, flickering blue which proclaimed how much she wanted to stay, here, with the rest of us, even as she left."\end{tabular}                                                                                                                                                                                                                                                                                                                                                                                                                                                                                                                                                                            \\ \cline{2-2} 
                    & \begin{tabular}[c]{@{}l@{}}The first person perspective in excerpt 1 is more interiorly concerned, elliptically mapping out the narrator's feelings toward \\ her mother's death: "That death itself would wait...what a ridiculous place to be when someone dies." Excerpt 2, though also \\ first person, allows more balance between interior and exterior insight, \underline{\color{ForestGreen}projecting the narrators feelings onto objects in her} \\ \underline{\color{ForestGreen}environment: "Bluer than the enamel-blue tulip-shaped pot that Bruce would get in the mail a few days later."}\end{tabular}                                                                              \\ \hline
\multirow{2}{*}{R3} & \begin{tabular}[c]{@{}l@{}}Both are fluent and coherent, but E2  is more effective. The \underline{\color{ForestGreen}language in E2 is more colorful and descriptive}. Compare for example \\ the description of the performance artist in E2: “He glanced across the room at a woman with strings of beads in her otherwise\\ naked Afro, then turned back to me . . . He sniffed at the air, feigned boredom and proceeded to chat up the Afro woman, who I \\ heard later was a performance artist.” Careful verb choices (glanced, sniffed, feigned) and \underline{\color{ForestGreen}concrete imagery (strings of beads)} \\ make the passage more effective than the stripped-down description in E1: “The agent shuffled past me to attend to a performance \\ artist so they could see what they could do for one another, before never speaking to each other again.”\end{tabular}                                                                                                                                                                                                                                                                                                                                                                                                                                                        \\ \cline{2-2} 
                    & \begin{tabular}[c]{@{}l@{}}\underline{\color{ForestGreen}The second excerpt uses livelier, more particular, even sensorial, language}; for example, the second excerpt states: "To be honest \\ I don’t think about race much, except when my liberal guilt compels me to consider it or when I read about some thirteen-year-old \\ kid being shot by some cop, and then I remember that though race is a construct without a basis in the natural world, those bullets \\ are quite real and deadly," which is rendered in the first excerpt as: "Race was a construct I didn't believe in and rarely dwelt on, \\ until those moments arose when guilt compelled me to see the "other side"of the matter. I saw its real-world consequences: police \\ brutality, discrimination, violence." The generic phrasing of "real-world consequences,"the abstract nouns like "brutality, \\ discrimination, violence," can't compare to the horrific and powerful vision of a "thirteen-year-old kid being shot by some cop," \\ and the feeling of bullets as "real and deadly." \underline{\color{ForestGreen}This specificity pervades the whole excerpt, from the description of the performance} \\ \underline{\color{ForestGreen}artist ("a woman with strings of beads in her otherwise naked Afro" in p2 vs simply "a performance artist" in p1).}\end{tabular}                 \\ \hline
\end{tabular}
\vspace{1ex}
\caption{\label{samereasoning}Similar lines of reasoning (R1, R2, R3) shared by expert writers on the same pair of excerpts during evaluation.}
\end{table*}

Experts on the other hand write much longer justifications (126 vs 79 words on average), and they more often quote or point to line-level evidence and use contrastive frames for their justification. Lay judges were instructed to do the same but we often find them not using such structure for their rationales. Experts talk more about the narrator's perspective, interiority, and how their POV shapes meaning. They reward concrete, discriminating observations about a character’s motives, relationships, and situation (not just ``it was emotional," but why and via which textual cues). They call out metaphor, symbol, motif, concrete images and how those devices carry the theme. They often praise earned tone while penalizing affected/performative voice and are more likely to mention syntax/pacing/arc, while simultaneously critiquing repetition or overwriting.

We observe that while judging writing quality is inherently rooted in subjectivity, our experts often use similar line of reasoning towards their preference. Experts often appreciate or critique the exact same trait that makes an excerpt better or weaker than the other. For example in Table \ref{samereasoning} both experts used the over expository nature of excerpt 2 as a common flaw that aimed to emulate Annie Ernaux's voice. Similarly for R2 both experts cited the same imagery \textit{Bluer than the enamel-blue tulip-shaped pot that Bruce would get in the mail a few days later} in the Cheryl Strayed emulation as a crucial element explaining their preference.

\subsection{After learning a preferred excerpt was written by AI, how do writers reconcile with others as well as their own judgments?\label{reconcile}}

\begin{table*}[!htbp]
\centering
\setlength{\tabcolsep}{8pt}
\renewcommand{\arraystretch}{1.05}
\small
\begin{tabular}{@{} p{0.19\textwidth} p{0.73\textwidth} c @{}}
\toprule
\textbf{Theme} & \textbf{Response} & \textbf{ID} \\
\midrule
Criteria Reframing &
\raggedright\q{This is part of what makes writing so interesting: there’s a huge level of subjectivity, as well as a sense of immediacy when you read something—especially short pieces deprived of context—where your initial reaction often takes over. Sometimes, you respond more positively to things that feel smoother or more polished at first glance, especially when comparing different articles. But I think it does speak to kind of the impressive nature of writing, to kind of be so broad and so subjective. And definitely it shows that} &
W1 \\
\addlinespace
Process Attribution &
\raggedright\q{When I'm working on the task, there are constraints on my time and on some level I'm viewing it as labor that I'm doing to get paid for a task... Whereas the LLM it's literally just detecting patterns and spitting out output.} &
W23 \\
\addlinespace
Technical Sensemaking &
\raggedright\q{I mean, apart from the initial emotional reaction that I just shared, I think one of the ways that I try to rationalize this as a writer is by understanding AI as something that thrives off of mass data. And so if you feed an immense amount of an original author's writing into a language model, it makes sense that it would be able to recognize the patterns in breadth and grammar and structure that I then praised.} &
W14 \\
\addlinespace
Expectation Violation &
\raggedright\q{A little upset. A little humbled, but mostly upset, I would say... I mean, it's deeply distressing.} &
W3 \\
\addlinespace
Capability Reassessment &
\raggedright\q{I do think that it means people should be much more skeptical of AI as a technology because I do think it's a lot more threatening than people anticipate it being, and I hear a lot it's just operating off of cliche. It's operating off of things that have already been written. I think seeing these excerpts partially—obviously this isn't like a novel or something—but at least on a line level it's a competent writer who is sometimes kind of funny and understands irony, which is not something that I thought that AI could do.} &
W15 \\
\bottomrule
\end{tabular}
\vspace{1ex}
\caption{Themes describing how participants made sense of their (as well as others') earlier preferences after discovering the content was AI-generated.}
\label{tab:reframing-table}
\end{table*}

We conducted thematic analysis following Braun and Clarke's  six-phase framework to identify patterns in responses provided by participants to our revelation and sensemaking \cite{braun2006using}. The authors of the paper independently conducted initial open coding of the transcripts, generating preliminary codes through line-by-line analysis. Through iterative discussion and comparison, we refined these codes and organized them into higher-level themes. This process therefore involved moving recursively between the coded data and emerging theme definitions, consistent with the reflexive thematic analysis approach \cite{braun2019reflecting}. Disagreements were resolved by consensus; unresolved alternatives were recorded as divergence memos and checked against additional data. Our analysis revealed five primary themes characterizing how participants made sense of others as well as their own earlier aesthetic preferences upon learning the content was AI-generated (Table \ref{tab:reframing-table}). Next, we describe these analytic themes and use brief data extracts to illustrate and support them.

\textsc{\textbf{Criteria Reframing}} describes instances in which participants recognize a mismatch between the criteria they actually used to evaluate the excerpts and the properties they report valuing in high-quality or ``authentic'' writing. These realizations prompt them to reconsider what, for them, constitutes ``good writing'' and whether their evaluative standards remain appropriate in light of AI's capabilities. For example, W10 noted having been ``looking primarily for just readability'' and worrying that they ``might [have] gone for the easier as opposed to the more unique or something newer,'' thereby acknowledging a tendency to privilege fluency over distinctiveness. Such reflections illustrate how participants move from treating their own criteria as largely implicit to subjecting them to more explicit scrutiny in response to encounters with AI-generated prose.

\textsc{\textbf{Process Attribution}} concerns how participants, once informed that a preferred excerpt was AI-generated, sought to account for differences between human and machine writing processes. Many emphasized that human writing is bound up with conscious effort, time pressure, and repeated cycles of drafting and revision, whereas AI systems can generate fluent text in a single pass. As W1 put it, one advantage of AI is that it can ``put out something that's super polished on the first attempt,'' in contrast to their own ``slower, more lengthy process that includes a lot of revision and going back.'' Such comments allow participants to reconcile AI's strong performance with their self-understanding as writers by locating the distinction not in surface polish, but in the temporal, iterative, and effortful character of human creative work.

\textsc{\textbf{Technical Sensemaking}} refers to a form of post-hoc explanation grounded in participants' (often partial) understanding of how large language models work. Rather than only expressing surprise or unease, participants attempted to reason through why the AI could produce persuasive prose by invoking ideas about training data, model scale, or computational resources. For instance, several described the model as drawing on ``mass data'' or ``pattern-matching'' across enormous corpora (see W14 in Table~\ref{tab:reframing-table}), using these accounts to relocate AI's advantage in technical infrastructure rather than in any intrinsically superior ``craft.'' In doing so, they turned an unsettling aesthetic outcome into something that could be explained in system-level terms.

\textsc{\textbf{Expectation Violation}} captures instances where the revelation that an excerpt was AI-generated conflicted with participants' prior beliefs about what AI can do or about their own ability to discern human from machine-written text, echoing classic work on expectancy violation \cite{burgoon1993interpersonal}. After the reveal, several writers described strong affective reactions. For example, W17 reported feeling ``pretty horrified'' and noted, ``I thought I had a better grasp of the ability to distinguish AI writing,'' describing a ``chilled feeling, especially as [a] writer,'' and finding it ``kind of scary'' that the excerpts were fully generated rather than merely AI-assisted. Such accounts indicate that the experimental feedback did more than surprise participants; it destabilized their sense of epistemic control over the boundary between human and AI-authored prose.

\textsc{\textbf{Capability Reassessment}} denotes moments in which participants substantially revise their prior beliefs about what AI can and cannot do in creative writing tasks. After the reveal, W13 remarked being ``just confused'' and that they ``didn't think that it had progressed to that level of sophistication,'' adding that they had ``always had faith that we would be able to discern something human from something technologically manufactured.'' Comments of this kind, echoed in other interviews (see W15 in Table~\ref{tab:reframing-table}), show participants recognizing that their earlier assumptions about AI’s limitations no longer hold and adjusting their sense of the technology’s reach in light of the excerpts they had just evaluated.

\subsection{How does choosing AI over human writing affect writers' professional identity and conceptions of ``good writing''?}

Similar to the process described in Section \ref{reconcile} we did thematic analysis following Braun and Clarke's  six-phase framework to identify patterns in responses to Identity Impact Questions. Our analysis revealed three primary themes characterizing how the discovery  of choosing AI over human writing destabilizes writers' understanding of what counts as "good writing" and their sense of professional identity.

\textsc{\textbf{Erosion of Aesthetic Confidence}} captures how the experiment unsettled writers’ prior sense that they could reliably recognize AI-generated text. Several participants reported being familiar with the ``feel'' of AI writing, echoing recent HCI work on readers’ sensitivity to LLM-produced prose \mbox{\cite{chakrabarty2024art,russell2025people}}, yet found that fine-tuned or carefully prompted outputs no longer fit that template. W15, who frequently edits high school essays, described themselves as ``pretty good at identifying AI-generated essays'' and as someone who would not typically ``enjoy AI generated writing,'' but nonetheless experienced the fine-tuned AI generations as ``surprising and frightening.'' Others drew out practical implications: W7 remarked that the study ``makes me feel like I'm going to have to run all my students' things through AI checkers now,'' while W2 admitted they had ``never thought a machine would be capable of style,'' finding that realization ``scary'' and ``frustrating.'' W1 similarly noted that, although they felt that they are able to recognize generic AI outputs that ``don't have any stylistic markers,'' the excerpts in this study ``definitely'' did not fit that pattern. Taken together, these reflections suggest that encounters with high-end AI emulations do not simply surprise writers; they actively erode confidence in their own aesthetic and diagnostic abilities.

\textsc{\textbf{Identity Crisis Around Expertise}} captures how the revelation that they had favored AI over human writing disrupted participants’ sense of themselves as skilled, distinctive practitioners. Several writers described the impact on their professional identity as both immediate and unsettling. W12, for example, reported returning to their own work to ask, "What is something that can't be mimicked?" and wondered, as large language models learn to reproduce features once thought "too intuitive or too emotionally complex to mimic," "What's the thing that is mine alone?" W9, who had felt confident in their emulation of an author they admire (Roxane Gay), confessed to having ``doubts about my own grasp over language'' and asked whether their writing would ``ever be to that level,'' resolving to attend to sentence structure ``more carefully and clearly.'' Others framed the episode as a kind of normative breach: as W20 put it, ``you want to always be picking the person,'' and they were ``more disappointed'' when they themselves selected the AI than when others did so. Taken together, these reactions suggest that encounters with highly capable AI do not merely prompt local revisions of taste, but pose a broader challenge to writers’ self-understanding as experts whose judgments and abilities mark them off from machines.

\textsc{\textbf{Redefining Writing's Purpose}} describes how several writers, when confronted with AI’s performance, shifted their account of writing’s value away from surface aesthetics toward process, intention, and human experience. W3, for instance, drew a distinction between using AI when ``you're writing to try to publish to make money'' and writing as ``a way of seeing or a way of thinking,'' arguing that AI ``eliminates what's process driven about writing and it eliminates what writing gives you.'' W17 similarly framed literature’s ``chief qualities'' as a dual track of providing ``emotional catharsis'' and producing ``novel thoughts, novel understandings of the world, of communities, of history.'' Others stressed innovation and breakage of convention, doubting that models trained on past work could genuinely extend the field; as W1 noted, creative writing may involve ``mixing up old tropes'' but also requires an ``element of insight where something completely new is produced.'' Across these comments, writers respond to AI’s evident stylistic competence by relocating the value of literary work in processes and effects they regard as closely tied to human perspective and agency.

\subsection{Writers' views on AI's implications for creative writing programs, careers, and professional institutions}

\subsubsection{\textbf{Impact of AI on the career of writers}} 
Writers almost unanimously expressed concern about AI's ability to emulate sophisticated literary voices. W8 articulated a core fear: when ``a large language model can develop or pantomime or mimic someone's voice to the point where it feels organic and the reader feels a sense of that author's essence being transmitted,'' this is ``incredibly worrying.'' For many, such capabilities function as a kind of ``canary in the coal mine'': in pursuit of commercial viability and audience appeal, writers produce more formulaic work that AI can readily emulate, and preserving distinctly human writing requires a renewed, radical commitment to craft. W20, for example, described AI as ``a threat to the ability for a writer to turn their craft into a livelihood,'' since writers are now ``competing with these AI versions of things that at least [are] more intelligible maybe for the common reader,'' but also framed their response as an obligation to ``put that energy back into writing something that isn't AI generated.''

Several participants also worried about how these dynamics would play out in an already precarious publishing landscape. Some suggested that the struggle and failure to get published in a market where AI-generated work finds commercial success might itself push writers toward using AI, even when they are uneasy about it. W3, speaking as both writer and reader, reported feeling ``more paranoid as a reader, more suspicious,'' and could ``see over time AI making it harder for people to pick up a book at random,'' especially from new authors, if trust and transparency around AI use remain limited. W6 voiced particular concern for debut and emerging authors, noting a broader pessimism about the industry's capacity to regulate AI: given that ``publishing is such a disaster,'' they were ``not convinced that it will be regulated well,'' and saw ``every chance that it could be difficult, especially for upcoming and debut authors if they're now competing with highly skilled Gen AI models in addition to competing with other, debut and emerging authors.''

At the same time, some writers identified reasons for cautious optimism. W2 emphasized the perceived importance of the author as a person and the relationship readers form with named writers, arguing that, even as AI eliminates certain jobs, they did not think ``creative writing is going to be one of those,'' in part because ``people like to have a name attached to things, especially when it's creative work,'' and ``don't think anybody wants'' a novel whose byline simply reads ``AI generated text.'' W10 similarly suggested that prevailing norms in literary communities may work in writers' favor: given how harshly authors caught plagiarizing are treated, they anticipated that undisclosed reliance on AI in creative writing would become a strong professional taboo, something people will ``definitely try to do and pass off, but [that] won't be widely accepted.'' Together, these responses portray a field in which AI is experienced both as a direct economic threat and as a catalyst for reaffirming professional norms around authorship, attribution, and the distinctiveness of human literary labor.

\subsubsection{\textbf{The Future of Creative Writing Programs}}
Literary scholar John Aldridge famously argued that MFA programs produce ``clonal fabrications of writers,'' implying a uniformity in style and output among graduates \mbox{\cite{so_piper_2016_mfa}}. Several participants explicitly echoed this concern. W9 remarked that ``the MFA \dots{} pushes you to write in this very specific and sort of odd way,'' noting that many students arrive as English majors who ``have the same canon'' and ``come in reading all the same books.'' W3 drew an even sharper parallel, suggesting that ``the homogenizing voice of AI is not different from the homogenizing impulse of American MFA programs,'' thereby linking institutional training and model training as two sources of stylistic convergence. These observations fed into a broader discussion of how creative writing programs might respond to AI. W19 argued that MFA programs will need to prioritize students who develop truly distinctive and exceptional styles, since AI excels at emulating established patterns, adding that ``we are already trained to think algorithmically about the books that we are writing \dots{} the market is inside our brains already.'' Others stressed the pedagogical implications for graduates who go on to teach. W10, for instance, pointed to a strong need for people who, by the end of their MFA, hope to become faculty to familiarize themselves with AI, suggesting that programs should include some form of training on how to distinguish AI-generated text from student work. Finally, some writers anticipated institutional diversification. One participant suggested that new MFA programs may explicitly incorporate AI into the curriculum, predicting a split between ``purist'' programs that keep AI at arm’s length—perhaps offering only one or two courses on ``writing in the age of AI'' as an object of reflection—and more precarious or visibility-seeking programs that market themselves as being at the forefront of AI-assisted creative practice (e.g., ``the first MFA with an AI partnership''). Across these views, AI is not simply treated as an external threat but as a force that may reconfigure how creative writing programs define their mission, select students, and design curricula.

\subsubsection{\textbf{How should literary platforms and publishing houses function in the wake of AI?}}
Several writers argued that, to maintain trust and integrity in the literary marketplace, it is crucial to disclose whether a work is written by a human or co-created with AI. W1 framed non-disclosure as a form of consumer deception, noting that readers purchase books with the expectation that ``there's another mind at work on the other side of this book,'' and warning that if this assumption no longer holds, it could ``devalue'' literary writing and generate broader ``distrust with the industry.'' Alongside disclosure, some participants stressed that copyright law should more clearly protect creative labor and clarify who counts as an author, with W9 bluntly suggesting that the appropriate response is to ``lobby for super strong copyright law.'' Others anticipated shifts in literary form and taste. W15 proposed that AI’s inability to have lived experience may produce stronger demand for autofictional and identitarian writing, wondering whether readers and publishers will increasingly value work that is explicitly rooted in an author’s own life because that is something ``ChatGPT can't do.'' At the same time, participants worried that, while self-published AI-generated ``slop'' is already flooding digital platforms such as Amazon, commercial pressures could tempt trade publishers to follow suit. W8 thought it plausible that ``one or two publishing houses [might] quietly'' deploy highly refined models for literary production, and argued that it would then fall to the literary community ``to sort of critique them into some form of shame or [into] rolling those back.'' Finally, several writers emphasized the need for stronger unions and collective organizations to respond to AI-driven changes in the industry. One participant observed that many US fiction writers come from relatively privileged backgrounds and therefore lack the economic pressure that has historically fostered solidarity in some genre communities, pointing to existing guild structures in science fiction and fantasy as a model. In their view, an appropriate response to AI would involve ``some level of solidarity and campaigning on the part of established writers,'' linking governance of AI use in publishing to broader questions of labor organization and bargaining power.

\section{Discussion}

\subsection{Writers have drawn inspiration from each other's style for centuries, so why should we hold AI to a different standard?}
Literary scholars argue that no text exists in isolation. The concept of \textit{intertextuality}, coined by Julia Kristeva, states that every text is a "mosaic of quotations" and is in dialogue with other texts \cite{orr2010intertextuality}. This means that writers are constantly, and often unconsciously, influenced by what they have read. In \textit{The Anxiety of Influence} \cite{bloom1997anxiety} eminent literary critic Harold Bloom pointed out how poets are hindered in their creative process by the ambiguous relationship they necessarily maintain with precursor poets. While admitting the influence of extraliterary experience on every poet, he argued that "the poet in a poet" is inspired to write by reading another poet's poetry and will tend to produce work that is in danger of being derivative of existing poetry. Such incidents have also existed in literary fiction as well. For instance Ian McEwan’s Atonement drew heavily on Lucilla Andrews’s memoir.\footnote{\url{https://www.theguardian.com/books/booksblog/2006/dec/08/plagarism}} Nobel laureate V.S. Naipaul, has spoken about his early efforts to write in the style of P.G. Wodehouse. Several major outlets have grouped Rachel Cusk with Knausgaard as catalysts of an autofiction wave—sometimes read as inspiration, sometimes as creating a template others follow.\footnote{\url{https://www.newstatesman.com/long-reads/2018/08/after-autofiction}} Bestselling author Sally Rooney has been hailed (and side-eyed) as ``Salinger for the Snapchat generation,''\footnote{\url{https://www.theguardian.com/culture/2017/dec/17/sally-rooney-author-on-my-radar-interview-mo-salah}} and has openly cited Franny and Zooey as an influence. Several critics have framed the recent vogue for vignette-driven, aphoristic fiction such as Jenny Offill’s second novel, ``Dept. of Speculation" as consciously working in  Lydia Davis's register. Considering these instances, AI emulation isn’t a new aesthetic phenomenon so much as an extreme of intertextuality \cite{brauneis2024copyright}. However what changes is the scale and cost. AI trained on millions of books can emulate the style for thousands of authors in seconds, with negligible marginal labor. Our experimental results support this, considering how the same GPT‑4o model can effortlessly emulate 30 authors and generate text on demand. This removal of time, apprenticeship, and friction that typically discipline influence also strips away the human filtering that yields an idiosyncratic voice. At population scale, this frictionless imitation risks saturation and substitution in a way no single writer’s influence ever could.

\subsection{How does this work contribute to the debate around Generative AI and Fair Use?}
A technologist, legal expert or user on the internet might think these results don't matter for fair use analysis since the outputs from the user preference don't directly copy the original works. While these outputs might match the quality and closely mirror the style of the source material, copyright law doesn't give authors exclusive rights to their writing style \cite{sobel2024elements}, and no one has a monopoly on producing high-quality literature. Some users have also asserted that while AI-generated outputs could serve as reasonable alternatives to an author's existing or future works, the same could be said for human-written works that draw inspiration from earlier texts. Particularly, in Bartz v. Anthropic copyright case Judge Alsup noted \textit{``But if someone were to read all the modern-day classics because of their exceptional expression, memorize them, and then emulate a blend of their best writing, would that violate the Copyright Act? Of course not}.'' This is fundamentally wrong because an important difference between human and AI-generated emulations is that humans read; AI systems copy. Unlike the semi-parametric memory of a billion-parameter LLM, human memory is not a verbatim storage device. Additionally, fine-tuning a model on 20 novels takes 3-4 hours whereas, an average human cannot read or consume information at this pace. We foresee in the future a situation could arise where a human can steer outputs from fine-tuned models to produce high-quality books/novellas. The US Copyright Office has already recognized how AI-generated derivative work may flood the market for the source works, resulting in ``market dilution,'' which is closely tied to Fair Use Factor 4.\footnote{The fourth fair use factor – often dubbed the single most important factor – instructs courts to inquire into the effect of the use upon the potential market for or value of the copyrighted work} In the \textit{Kadrey v. Meta}, copyright case, Judge Chhabria granted summary judgment to Meta, but accepted the theory of market dilution. However Judge Chhabria suggested in his opinion that some kinds of books may be less subject to substitution than others. In his words ``\textit{It seems unlikely, for instance, that AI-generated books would meaningfully siphon sales away from well-known authors who sell books to people looking for books by those particular authors.''} Our findings (Section 4.1) contradict Judge Chhabria's assumption that authors with unique writing styles are naturally protected from being replaced. Without disclosure if readers actually enjoy AI-generated imitations of well-known authors—particularly those valued for their distinctive voices—then all authors could face significant competition, especially from AI systems fine-tuned specifically on their work. Since producing AI-generated content costs much less than paying human writers, this makes it even more likely that the original authors' market share could be eroded. We examine the implications of AI model fine-tuning on Fair Use Factor 4 by analyzing a larger dataset in our contemporaneous work~\cite{chakrabarty2025readersPreferCopyrightBooks}.

\subsection{Can we regulate hidden AI authorship?}
Creators may be motivated to hide the involvement of generative AI in creating a work, since acknowledging it could undermine their chances of obtaining copyright protection and profiting from the work. This brings in interesting questions around whether we can regulate such practices. First and foremost there is a dire need to update beliefs around AI detection. Some commercial AI detectors aren't accurate, but that should not lead to the discourse that AI detectors don't work. As a matter of fact text generated by prompting AI can be detected accurately as shown by recent work \cite{russell2025people, jabarian_imas_2025_artificial}. Text generated from fine-tuned models however poses a separate kind of risk as they evade standard AI detectors, due to distributional similarity to human-written text. Here, watermarking remains a potential solution, especially post-hoc watermarking that evades paraphrase attacks \cite{chang2024postmark,xu2024robust}. However at the crux of all these technical solutions, there needs to be legal restriction such as the EU AI Act’s Article 50, that obligates providers of systems generating synthetic audio/image/video/text to ensure outputs are marked and detectable. Publishing houses or Online Platforms such as Kindle could have certification boards that conduct proper investigation and commit to certifying how content was not produced by generative AI. As pointed by \citet{noti2025regulating}, the FTC can also treat undisclosed AI authorship (when material to consumers) as a misleading omission and bring enforcement actions—even without a blanket federal disclosure law. We discuss AI detectability and how removing ``AI quirks'' changes preference patterns via 
additional analyses using a larger dataset in a related paper~\cite{chakrabarty2025readersPreferCopyrightBooks}.

\section{Limitations and Future Work}
We tried our best to design a fair experiment, but we were constrained by some practical limitations. For instance a larger pool of experts would add more credibility to our study, but it was challenging to recruit more writers, given how our affiliation as AI researchers prevented writers from working with us. Our recruitment was mostly restricted to American creative writing programs and further study needs to be done across creative writing programs outside the US. While our pre-selected pool of 50 writers consisted of some writers who do not write in English, our experiments on style/voice emulation were done based on their English translation. Creative writing often depends on intrinsic motivation. While we offered MFA students a lucrative rate for writing the excerpts, it's unclear if monetary incentives actually enhanced their creative output, since intrinsic motivation typically drives the best artistic work. Additionally there is a possibility that MFA trained expert writers are not very good at mimicking other writers. We acknowledge MFA trained expert writers are not monolithic and themselves have different skillsets and styles. Last but not least our experiments were conducted at a shorter excerpt level and conclusions cannot be drawn for long-form text. In its current form, AI is unable to generate long-form text that's thematically coherent unlike humans. While we foresee a situation where humans can collaborate with a fine-tuned AI model to create competing long-form works, experimental evidence is required to make any broader claims.

\section{Conclusion}

 In this work we conduct a controlled experiment with professionally trained human writers and LLMs emulating critically acclaimed authors. We find that lay readers who represent a big chunk of the consumer base preferred AI  over human writing. At the same time while expert readers strongly preferred human writing over AI when generated via in-context prompting, this preference reversed drastically with fine-tuning. Our results challenge fundamental assumptions about whether good writing can be generative. Based on debrief interviews we find that discovering their preference for AI writing triggered profound responses among writers. Many  writers attempted to redefine writing's value away from aesthetic quality toward process and intention, suggesting that if AI can match human writing style/voice, perhaps writing's worth lies in the human experience of creation rather than the output itself. The implications of our experiments go beyond individual writers and are applicable to the entire literary ecosystem. Our findings show that fine-tuning on copyrighted data leads to a form of style extraction that has a strong potential for labor market dilution. While copyright law doesn't protect style as such, our results show that AI trained on authors' complete works can produce text that readers including experts find superior to human writing. As AI gets more and more capable, the question is no longer whether it can write well, but rather how we navigate a world where it can. 
 
\bibliographystyle{ACM-Reference-Format}
\bibliography{bibliography}

@article{menand2009show,
  author  = {Louis Menand},
  title   = {Show or Tell},
  journal = {The New Yorker},
  year    = {2009},
  month   = jun,
  note    = {A Critic at Large; June 8, 2009 issue},
  url     = {https://www.newyorker.com/magazine/2009/06/08/show-or-tell}
}

@book{mcgurl2011program,
  title={The program era: Postwar fiction and the rise of creative writing},
  author={McGurl, Mark},
  year={2011},
  publisher={Harvard University Press}
}

@article{doshi2024generative,
  title={Generative AI enhances individual creativity but reduces the collective diversity of novel content},
  author={Doshi, Anil R and Hauser, Oliver P},
  journal={Science Advances},
  volume={10},
  number={28},
  pages={eadn5290},
  year={2024},
  publisher={American Association for the Advancement of Science}
}

@article{ling2025underreporting,
  title={Underreporting of AI use: The role of social desirability bias},
  author={Ling, Yier and Imas, Alex},
  journal={Available at SSRN},
  year={2025}
}

@inproceedings{li-etal-2024-disclosure,
    title = "How Does the Disclosure of {AI} Assistance Affect the Perceptions of Writing?",
    author = "Li, Zhuoyan  and
      Liang, Chen  and
      Peng, Jing  and
      Yin, Ming",
    editor = "Al-Onaizan, Yaser  and
      Bansal, Mohit  and
      Chen, Yun-Nung",
    booktitle = "Proceedings of the 2024 Conference on Empirical Methods in Natural Language Processing",
    month = nov,
    year = "2024",
    address = "Miami, Florida, USA",
    publisher = "Association for Computational Linguistics",
    url = "https://aclanthology.org/2024.emnlp-main.279/",
    doi = "10.18653/v1/2024.emnlp-main.279",
    pages = "4849--4868"
}

@article{denzin2001reflexive,
  title={The reflexive interview and a performative social science},
  author={Denzin, Norman K},
  journal={Qualitative research},
  volume={1},
  number={1},
  pages={23--46},
  year={2001},
  publisher={Sage Publications Sage CA: Thousand Oaks, CA}
}

@article{finlay2002negotiating,
  title={Negotiating the swamp: the opportunity and challenge of reflexivity in research practice},
  author={Finlay, Linda},
  journal={Qualitative research},
  volume={2},
  number={2},
  pages={209--230},
  year={2002},
  publisher={Sage Publications Sage CA: Thousand Oaks, CA}
}

@article{birt2016member,
  title={Member checking: a tool to enhance trustworthiness or merely a nod to validation?},
  author={Birt, Linda and Scott, Suzanne and Cavers, Debbie and Campbell, Christine and Walter, Fiona},
  journal={Qualitative health research},
  volume={26},
  number={13},
  pages={1802--1811},
  year={2016},
  publisher={Sage Publications Sage CA: Los Angeles, CA}
}

@book{lincoln1985naturalistic,
  title={Naturalistic inquiry},
  author={Lincoln, Yvonna S},
  volume={75},
  year={1985},
  publisher={sage}
}

@book{patton2014qualitative,
  title={Qualitative research \& evaluation methods: Integrating theory and practice},
  author={Patton, Michael Quinn},
  year={2014},
  publisher={Sage publications}
}

@article{li2023self,
  title={Self-alignment with instruction backtranslation},
  author={Li, Xian and Yu, Ping and Zhou, Chunting and Schick, Timo and Levy, Omer and Zettlemoyer, Luke and Weston, Jason and Lewis, Mike},
  journal={arXiv preprint arXiv:2308.06259},
  year={2023}
}

@article{chiang_why_2024,
  author       = {Chiang, Ted},
  title        = {Why A.{I}. Isn’t Going to Make Art},
  journal      = {The New Yorker},
  year         = {2024},
  month        = {August},
  day          = {31},
  note         = {The Weekend Essay},
  url          = {https://www.newyorker.com/culture/the-weekend-essay/why-ai-isnt-going-to-make-art},
  accessdate   = {2025-06-21},
}

@inproceedings{sarkar2025ai,
  title={AI Could Have Written This: Birth of a Classist Slur in Knowledge Work},
  author={Sarkar, Advait},
  booktitle={Proceedings of the Extended Abstracts of the CHI Conference on Human Factors in Computing Systems},
  pages={1--12},
  year={2025}
}

@inproceedings{anderson2024homogenization,
  title={Homogenization effects of large language models on human creative ideation},
  author={Anderson, Barrett R and Shah, Jash Hemant and Kreminski, Max},
  booktitle={Proceedings of the 16th conference on creativity \& cognition},
  pages={413--425},
  year={2024}
}

@inproceedings{mcdonnell2016relevant,
  title={Why is that relevant? collecting annotator rationales for relevance judgments},
  author={McDonnell, Tyler and Lease, Matthew and Kutlu, Mucahid and Elsayed, Tamer},
  booktitle={Proceedings of the AAAI Conference on Human Computation and Crowdsourcing},
  volume={4},
  pages={139--148},
  year={2016}
}

@article{noti2025regulating,
  title={Regulating Hidden AI Authorship},
  author={Noti-Victor, Jacob},
  journal={Va. L. Rev.},
  volume={111},
  pages={139},
  year={2025},
  publisher={HeinOnline}
}

@article{Tangermann2025,
  author       = {Tangermann, Victor},
  title        = {Readers Annoyed When Fantasy Novel Accidentally Leaves AI Prompt in Published Version, Showing Request to Copy Another Writer's Style},
  journal      = {Futurism},
  year         = {2025},
  month        = {May},
  day          = {23},
  url          = {https://futurism.com/fantasy-novel-ai-prompt-copy-style},
  note         = {Accessed: 2025-06-21},
}

@inproceedings{gero2025creative,
  title={Creative Writers' Attitudes on Writing as Training Data for Large Language Models},
  author={Gero, Katy Ilonka and Desai, Meera and Schnitzler, Carly and Eom, Nayun and Cushman, Jack and Glassman, Elena L},
  booktitle={Proceedings of the 2025 CHI Conference on Human Factors in Computing Systems},
  pages={1--16},
  year={2025}
}

@inproceedings{chakrabarty2025can,
  title={Can {AI} writing be salvaged? {M}itigating idiosyncrasies and improving human-ai alignment in the writing process through edits},
  author={Chakrabarty, Tuhin and Laban, Philippe and Wu, Chien-Sheng},
  booktitle={Proceedings of the 2025 CHI Conference on Human Factors in Computing Systems},
  pages={1--33},
  year={2025}
}

@article{burgoon1993interpersonal,
  title={Interpersonal expectations, expectancy violations, and emotional communication},
  author={Burgoon, Judee K},
  journal={Journal of language and social psychology},
  volume={12},
  number={1-2},
  pages={30--48},
  year={1993},
  publisher={Sage Publications Sage CA: Thousand Oaks, CA}
}

@article{xu2024robust,
  title={Robust Multi-bit Text Watermark with LLM-based Paraphrasers},
  author={Xu, Xiaojun and Jia, Jinghan and Yao, Yuanshun and Liu, Yang and Li, Hang},
  journal={arXiv preprint arXiv:2412.03123},
  year={2024}
}

@article{chang2024postmark,
  title={Postmark: A robust blackbox watermark for large language models},
  author={Chang, Yapei and Krishna, Kalpesh and Houmansadr, Amir and Wieting, John and Iyyer, Mohit},
  journal={arXiv preprint arXiv:2406.14517},
  year={2024}
}

@article{russell2025people,
  title={People who frequently use ChatGPT for writing tasks are accurate and robust detectors of AI-generated text},
  author={Russell, Jenna and Karpinska, Marzena and Iyyer, Mohit},
  journal={arXiv preprint arXiv:2501.15654},
  year={2025}
}

@misc{jabarian_imas_2025_artificial,
  author    = {Jabarian, Brian and Imas, Alex},
  title     = {Artificial Writing and Automated Detection},
  year      = {2025},
  month     = aug,
  note      = {SSRN working paper, Abstract ID 5407424},
  url       = {https://ssrn.com/abstract=5407424},
  urldate   = {2025-08-31}
}

@article{so_piper_2016_mfa,
  author       = {So, Richard Jean and Piper, Andrew},
  title        = {How Has the MFA Changed the Contemporary Novel?},
  journal      = {The Atlantic},
  year         = {2016},
  month        = mar,
  day          = {6},
  url          = {https://www.theatlantic.com/entertainment/archive/2016/03/mfa-creative-writing/462483/},
}

@inproceedings{gero2023social,
  title={Social dynamics of AI support in creative writing},
  author={Gero, Katy Ilonka and Long, Tao and Chilton, Lydia B},
  booktitle={Proceedings of the 2023 CHI conference on human factors in computing systems},
  pages={1--15},
  year={2023}
}

@inproceedings{zhao2025making,
  title={Making the Write Connections: Linking Writing Support Tools with Writer Needs},
  author={Zhao, Zixin and Masson, Damien and Kim, Young-Ho and Penn, Gerald and Chevalier, Fanny},
  booktitle={Proceedings of the 2025 CHI Conference on Human Factors in Computing Systems},
  pages={1--21},
  year={2025}
}

@inproceedings{10.1145/3715336.3735848,
author = {Wise, Talia and Yang, Yuewen and Shim, Ryun and Chang, Kevin Chuan-Kai and Oden Choi, Judeth and Yang, Qian},
title = {Investigating How Emerging Adults Explore Identity through Writing: Opportunities for AI Writing Assistants to Help},
year = {2025},
isbn = {9798400714856},
publisher = {Association for Computing Machinery},
address = {New York, NY, USA},
url = {https://doi.org/10.1145/3715336.3735848},
doi = {10.1145/3715336.3735848},
booktitle = {Proceedings of the 2025 ACM Designing Interactive Systems Conference},
pages = {2270–2282},
numpages = {13},
location = {
},
series = {DIS '25}
}

@inproceedings{guo2025pen,
  title={From pen to prompt: how creative writers integrate AI into their writing practice},
  author={Guo, Alicia and Sathyanarayanan, Shreya and Wang, Leijie and Heer, Jeffrey and Zhang, Amy X},
  booktitle={Proceedings of the 2025 Conference on Creativity and Cognition},
  pages={527--545},
  year={2025}
}

@inproceedings{dhillon2024shaping,
  title={Shaping human-AI collaboration: Varied scaffolding levels in co-writing with language models},
  author={Dhillon, Paramveer S and Molaei, Somayeh and Li, Jiaqi and Golub, Maximilian and Zheng, Shaochun and Robert, Lionel Peter},
  booktitle={Proceedings of the 2024 CHI Conference on Human Factors in Computing Systems},
  pages={1--18},
  year={2024}
}

@inproceedings{calderwood2025phraselette,
  title={Phraselette: A Poet’s Procedural Palette},
  author={Calderwood, Alex and Chung, John Joon Young and Sun, Yuqian and Roemmele, Melissa and Kreminski, Max},
  booktitle={Proceedings of the 2025 ACM Designing Interactive Systems Conference},
  pages={2701--2717},
  year={2025}
}

@article{draxler2024ai,
  title={The AI ghostwriter effect: When users do not perceive ownership of AI-generated text but self-declare as authors},
  author={Draxler, Fiona and Werner, Anna and Lehmann, Florian and Hoppe, Matthias and Schmidt, Albrecht and Buschek, Daniel and Welsch, Robin},
  journal={ACM Transactions on Computer-Human Interaction},
  volume={31},
  number={2},
  pages={1--40},
  year={2024},
  publisher={ACM New York, NY}
}

@article{hwang202580,
  title={'It was 80\% me, 20\% AI': Seeking Authenticity in Co-Writing with Large Language Models},
  author={Hwang, Angel Hsing-Chi and Liao, Q Vera and Blodgett, Su Lin and Olteanu, Alexandra and Trischler, Adam},
  journal={Proceedings of the ACM on Human-Computer Interaction},
  volume={9},
  number={2},
  pages={1--41},
  year={2025},
  publisher={ACM New York, NY, USA}
}

@article{ippolito2022creative,
  title={Creative Writing with an AI-Powered Writing Assistant: Perspectives from Professional Writers},
  author={Ippolito, Daphne and Yuan, Ann and Coenen, Andy and Burnam, Sehmon},
  journal={arXiv preprint arXiv:2211.05030},
  year={2022}
}

@article{liang2025quantifying,
  title={Quantifying large language model usage in scientific papers},
  author={Liang, Weixin and Zhang, Yaohui and Wu, Zhengxuan and Lepp, Haley and Ji, Wenlong and Zhao, Xuandong and Cao, Hancheng and Liu, Sheng and He, Siyu and Huang, Zhi and others},
  journal={Nature Human Behaviour},
  pages={1--11},
  year={2025},
  publisher={Nature Publishing Group UK London}
}

@inproceedings{yuan2022wordcraft,
  title={Wordcraft: story writing with large language models},
  author={Yuan, Ann and Coenen, Andy and Reif, Emily and Ippolito, Daphne},
  booktitle={Proceedings of the 27th International Conference on Intelligent User Interfaces},
  pages={841--852},
  year={2022}
}

@inproceedings{mirowski2024robot,
  title={A Robot Walks into a Bar: Can Language Models Serve as Creativity SupportTools for Comedy? An Evaluation of LLMs’ Humour Alignment with Comedians},
  author={Mirowski, Piotr and Love, Juliette and Mathewson, Kory and Mohamed, Shakir},
  booktitle={The 2024 ACM Conference on Fairness, Accountability, and Transparency},
  pages={1622--1636},
  year={2024}
}

@inproceedings{mirowski2023cowriting,
author = {Mirowski, Piotr and Mathewson, Kory W. and Pittman, Jaylen and Evans, Richard},
title = {Co-Writing Screenplays and Theatre Scripts with Language Models: Evaluation by Industry Professionals},
year = {2023},
isbn = {9781450394215},
publisher = {Association for Computing Machinery},
address = {New York, NY, USA},
url = {https://doi.org/10.1145/3544548.3581225},
doi = {10.1145/3544548.3581225},
booktitle = {Proceedings of the 2023 CHI Conference on Human Factors in Computing Systems},
articleno = {355},
numpages = {34},
location = {Hamburg, Germany},
series = {CHI '23}
}

@inproceedings{10.1145/3635636.3656201,
author = {Chakrabarty, Tuhin and Padmakumar, Vishakh and Brahman, Faeze and Muresan, Smaranda},
title = {Creativity Support in the Age of Large Language Models: An Empirical Study Involving Professional Writers},
year = {2024},
isbn = {9798400704857},
publisher = {Association for Computing Machinery},
address = {New York, NY, USA},
url = {https://doi.org/10.1145/3635636.3656201},
doi = {10.1145/3635636.3656201},
booktitle = {Proceedings of the 16th Conference on Creativity \& Cognition},
pages = {132–155},
numpages = {24},
location = {Chicago, IL, USA},
series = {C \&C '24}
}

@book{bloom1997anxiety,
  title={The anxiety of influence: A theory of poetry},
  author={Bloom, Harold},
  year={1997},
  publisher={Oxford University Press}
}

@article{nugent2020imitateSaunders,
  author  = {Nugent, Benjamin},
  title   = {How to Imitate George Saunders},
  journal = {The Paris Review},
  year    = {2020},
  month   = {1},
  day     = {6},
  url     = {https://www.theparisreview.org/blog/2020/01/06/how-to-imitate-george-saunders/},
  note    = {Accessed: 2025-11-26}
}

@article{vanderark2018saundersLearningWrite,
  author  = {Vander Ark, Tom},
  title   = {George Saunders On Learning to Write--and Writing to Learn},
  journal = {Getting Smart},
  year    = {2018},
  month   = {2},
  day     = {15},
  url     = {https://www.gettingsmart.com/2018/02/15/george-saunders-learning-write-writing-learn/},
  note    = {Accessed: 2025-11-26}
}

@online{jonesaiknockoffs2025,
  author       = {CT Jones},
  title        = {Amazon Is the World’s Biggest Online Book Marketplace. It’s Filled With AI Knockoffs},
  subtitle     = {Authors say Amazon’s knockoff book problem is leaving them frustrated—and making the internet worse in the process},
  date         = {2025-10-27},
  url          = {https://www.rollingstone.com/culture/culture-features/amazon-ai-book-knockoffs-1235450690/},
  urldate      = {2025-10-29},
  organization = {Rolling Stone},
  publisher    = {Penske Media Corporation},
  langid       = {english},
  note         = {Culture Feature}
}

@online{knibbsscammyaibooks2024,
  author       = {Kate Knibbs},
  title        = {Scammy AI-Generated Book Rewrites Are Flooding Amazon},
  date         = {2024-01-10},
  url          = {https://www.wired.com/story/scammy-ai-generated-books-flooding-amazon/},
  urldate      = {2025-10-29},
  organization = {WIRED},
  publisher    = {Condé Nast},
  langid       = {english},
  note         = {Business}
}

@online{varaairomancefactory2025,
  author       = {Vauhini Vara},
  title        = {The A.I. Romance Factory},
  subtitle     = {Genre fiction publisher Inkitt has influential backers and a vision for infinitely customizable A.I.-driven content},
  date         = {2025-04-07},
  url          = {https://www.bloomberg.com/features/2025-ai-romance-factory/},
  urldate      = {2025-10-29},
  organization = {Bloomberg Businessweek},
  publisher    = {Bloomberg L.P.},
  langid       = {english},
  note         = {Feature}
}

@article{corbett1971theory,
  title={The theory and practice of imitation in classical rhetoric},
  author={Corbett, Edward PJ},
  journal={College Composition \& Communication},
  volume={22},
  number={3},
  pages={243--250},
  year={1971},
  publisher={NCTE}
}

@article{orr2010intertextuality,
  title={Intertextuality},
  author={Orr, Mary},
  journal={The encyclopedia of literary and cultural theory},
  year={2010},
  publisher={Wiley Online Library}
}

@article{sobel2024elements,
  title={Elements of style: copyright, similarity, and generative AI},
  author={Sobel, Benjamin},
  journal={Harvard Journal of Law \& Technology, Forthcoming},
  volume={38},
  year={2024}
}

@article{erickson2024ai,
  title={AI and work in the creative industries: digital continuity or discontinuity?},
  author={Erickson, Kristofer},
  journal={Creative Industries Journal},
  pages={1--21},
  year={2024},
  publisher={Taylor \& Francis}
}

@inproceedings{porquet2025copying,
  title={Copying style, Extracting value: Illustrators' Perception of AI Style Transfer and its Impact on Creative Labor},
  author={Porquet, Julien and Wang, Sitong and Chilton, Lydia B},
  booktitle={Proceedings of the 2025 CHI Conference on Human Factors in Computing Systems},
  pages={1--16},
  year={2025}
}

@inproceedings{wadinambiarachchi2024effects,
  title={The effects of generative ai on design fixation and divergent thinking},
  author={Wadinambiarachchi, Samangi and Kelly, Ryan M and Pareek, Saumya and Zhou, Qiushi and Velloso, Eduardo},
  booktitle={Proceedings of the 2024 CHI Conference on Human Factors in Computing Systems},
  pages={1--18},
  year={2024}
}

@inproceedings{lin-2004-rouge,
    title = "{ROUGE}: A Package for Automatic Evaluation of Summaries",
    author = "Lin, Chin-Yew",
    booktitle = "Text Summarization Branches Out",
    month = jul,
    year = "2004",
    address = "Barcelona, Spain",
    publisher = "Association for Computational Linguistics",
    url = "https://aclanthology.org/W04-1013/",
    pages = "74--81"
}

@inproceedings{kyi2025governance,
  title={Governance of Generative AI in Creative Work: Consent, Credit, Compensation, and Beyond},
  author={Kyi, Lin and Mahuli, Amruta and Silberman, M Six and Binns, Reuben and Zhao, Jun and Biega, Asia J},
  booktitle={Proceedings of the 2025 CHI Conference on Human Factors in Computing Systems},
  pages={1--16},
  year={2025}
}

@inproceedings{he2025contributions,
  title={Which contributions deserve credit? perceptions of attribution in human-ai co-creation},
  author={He, Jessica and Houde, Stephanie and Weisz, Justin D},
  booktitle={Proceedings of the 2025 CHI Conference on Human Factors in Computing Systems},
  pages={1--18},
  year={2025}
}

@inproceedings{lima2025public,
  title={Public Opinions About Copyright for AI-Generated Art: The Role of Egocentricity, Competition, and Experience},
  author={Lima, Gabriel and Grgi{\'c}-Hla{\v{c}}a, Nina and Redmiles, Elissa M},
  booktitle={Proceedings of the 2025 CHI Conference on Human Factors in Computing Systems},
  pages={1--32},
  year={2025}
}

@techreport{usdcc_ndca_bartz_anthropic_2025,
  author       = {{United States District Court for the Northern District of California}},
  title        = {Order on Fair Use},
  institution  = {United States District Court, Northern District of California},
  type         = {Court Opinion},
  number       = {No. C 24-05417 WHA, Doc. 231},
  date         = {2025-06-23},
  url          = {https://storage.courtlistener.com/recap/gov.uscourts.cand.434709/gov.uscourts.cand.434709.231.0.pdf},
  note         = {Accessed: July 2, 2025}
}

@article{chakrabarty2025readersPreferCopyrightBooks,
  title   = {Readers Prefer Outputs of AI Trained on Copyrighted Books over Expert Human Writers},
  author  = {Chakrabarty, Tuhin and Ginsburg, Jane C. and Dhillon, Paramveer S.},
  year    = {2025},
  journal = {SSRN},
  note    = {Columbia Public Law Research Paper No. 5606570},
  doi     = {10.2139/ssrn.5606570}
}

@article{zhang2026owns,
  title={Who Owns the Text? Design Patterns for Preserving Authorship in AI-Assisted Writing},
  author={Zhang, Bohan and Bu, Chengke and Dhillon, Paramveer S},
  journal={arXiv preprint arXiv:2601.10236},
  year={2026}
}

@article{10.1093/jiplp/jpaf046,
    author = {Ginsburg, Jane C},
    title = {AI inputs, fair use and the US Copyright Office Report},
    journal = {Journal of Intellectual Property Law and Practice},
    volume = {20},
    number = {8},
    pages = {521-522},
    year = {2025},
    month = {08},
    issn = {1747-1532},
    doi = {10.1093/jiplp/jpaf046},
    url = {https://academic.oup.com/jiplp/article-pdf/20/8/521/63928340/jpaf046.pdf},
}

@article{brauneis2024copyright,
  title={Copyright and the training of human authors and generative machines},
  author={Brauneis, Robert},
  journal={Colum. JL \& Arts},
  volume={48},
  pages={1},
  year={2024},
  publisher={HeinOnline}
}

@article{samuelson2023generative,
  title={Generative AI meets copyright},
  author={Samuelson, Pamela},
  journal={Science},
  volume={381},
  number={6654},
  pages={158--161},
  year={2023},
  publisher={American Association for the Advancement of Science}
}

@misc{hansen_bartz_anthropic_2025,
  author       = {Hansen, Dave},
  title        = {Bartz v. Anthropic: A Preliminary Look at What LibGen Books May Be Included in the Class Action},
  year         = {2025},
  month        = sep,
  howpublished = {Authors Alliance},
  note         = {Blog post},
  url          = {https://www.authorsalliance.org/2025/09/05/bartz-v-anthropic-a-preliminary-look-at-what-libgen-books-may-be-included-in-the-class-action/},
  urldate      = {2025-09-06}
}

@inproceedings{gero2022sparks,
  title={Sparks: Inspiration for science writing using language models},
  author={Gero, Katy Ilonka and Liu, Vivian and Chilton, Lydia},
  booktitle={Proceedings of the 2022 ACM Designing Interactive Systems Conference},
  pages={1002--1019},
  year={2022}
}

@article{touvron2023llama,
  title={Llama: Open and efficient foundation language models},
  author={Touvron, Hugo and Lavril, Thibaut and Izacard, Gautier and Martinet, Xavier and Lachaux, Marie-Anne and Lacroix, Timoth{\'e}e and Rozi{\`e}re, Baptiste and Goyal, Naman and Hambro, Eric and Azhar, Faisal and others},
  journal={arXiv preprint arXiv:2302.13971},
  year={2023}
}

@article{brown2020language,
  title={Language models are few-shot learners},
  author={Brown, Tom and Mann, Benjamin and Ryder, Nick and Subbiah, Melanie and Kaplan, Jared D and Dhariwal, Prafulla and Neelakantan, Arvind and Shyam, Pranav and Sastry, Girish and Askell, Amanda and others},
  journal={Advances in neural information processing systems},
  volume={33},
  pages={1877--1901},
  year={2020}
}

@article{kobak2025delving,
  title={Delving into LLM-assisted writing in biomedical publications through excess vocabulary},
  author={Kobak, Dmitry and Gonz{\'a}lez-M{\'a}rquez, Rita and Horv{\'a}t, Em{\H{o}}ke-{\'A}gnes and Lause, Jan},
  journal={Science Advances},
  volume={11},
  number={27},
  pages={eadt3813},
  year={2025},
  publisher={American Association for the Advancement of Science}
}

@article{liang2024mapping,
  title={Mapping the increasing use of LLMs in scientific papers},
  author={Liang, Weixin and Zhang, Yaohui and Wu, Zhengxuan and Lepp, Haley and Ji, Wenlong and Zhao, Xuandong and Cao, Hancheng and Liu, Sheng and He, Siyu and Huang, Zhi and others},
  journal={arXiv preprint arXiv:2404.01268},
  year={2024}
}

@article{liang2025widespread,
  title={The Widespread Adoption of Large Language Model-Assisted Writing Across Society},
  author={Liang, Weixin and Zhang, Yaohui and Codreanu, Mihai and Wang, Jiayu and Cao, Hancheng and Zou, James},
  journal={arXiv preprint arXiv:2502.09747},
  year={2025}
}

@inproceedings{lee2024design,
  title={A Design Space for Intelligent and Interactive Writing Assistants},
  author={Lee, Mina and Gero, Katy Ilonka and Chung, John Joon Young and Shum, Simon Buckingham and Raheja, Vipul and Shen, Hua and Venugopalan, Subhashini and Wambsganss, Thiemo and Zhou, David and Alghamdi, Emad A and others},
  booktitle={Proceedings of the CHI Conference on Human Factors in Computing Systems},
  pages={1--35},
  year={2024}
}

@article{braun2019reflecting,
  title={Reflecting on reflexive thematic analysis},
  author={Braun, Virginia and Clarke, Victoria},
  journal={Qualitative research in sport, exercise and health},
  volume={11},
  number={4},
  pages={589--597},
  year={2019},
  publisher={Taylor \& Francis}
}

@inproceedings{chakrabarty2024art,
  title={Art or artifice? large language models and the false promise of creativity},
  author={Chakrabarty, Tuhin and Laban, Philippe and Agarwal, Divyansh and Muresan, Smaranda and Wu, Chien-Sheng},
  booktitle={Proceedings of the 2024 CHI Conference on Human Factors in Computing Systems},
  pages={1--34},
  year={2024}
}

@article{braun2006using,
  title={Using thematic analysis in psychology},
  author={Braun, Virginia and Clarke, Victoria},
  journal={Qualitative research in psychology},
  volume={3},
  number={2},
  pages={77--101},
  year={2006},
  publisher={Taylor \& Francis}
}

\appendix 
\section{Appendix}\label{sec:appendix}

\begin{figure*}[!htbp]
    \centering
    \includegraphics[width=\textwidth]{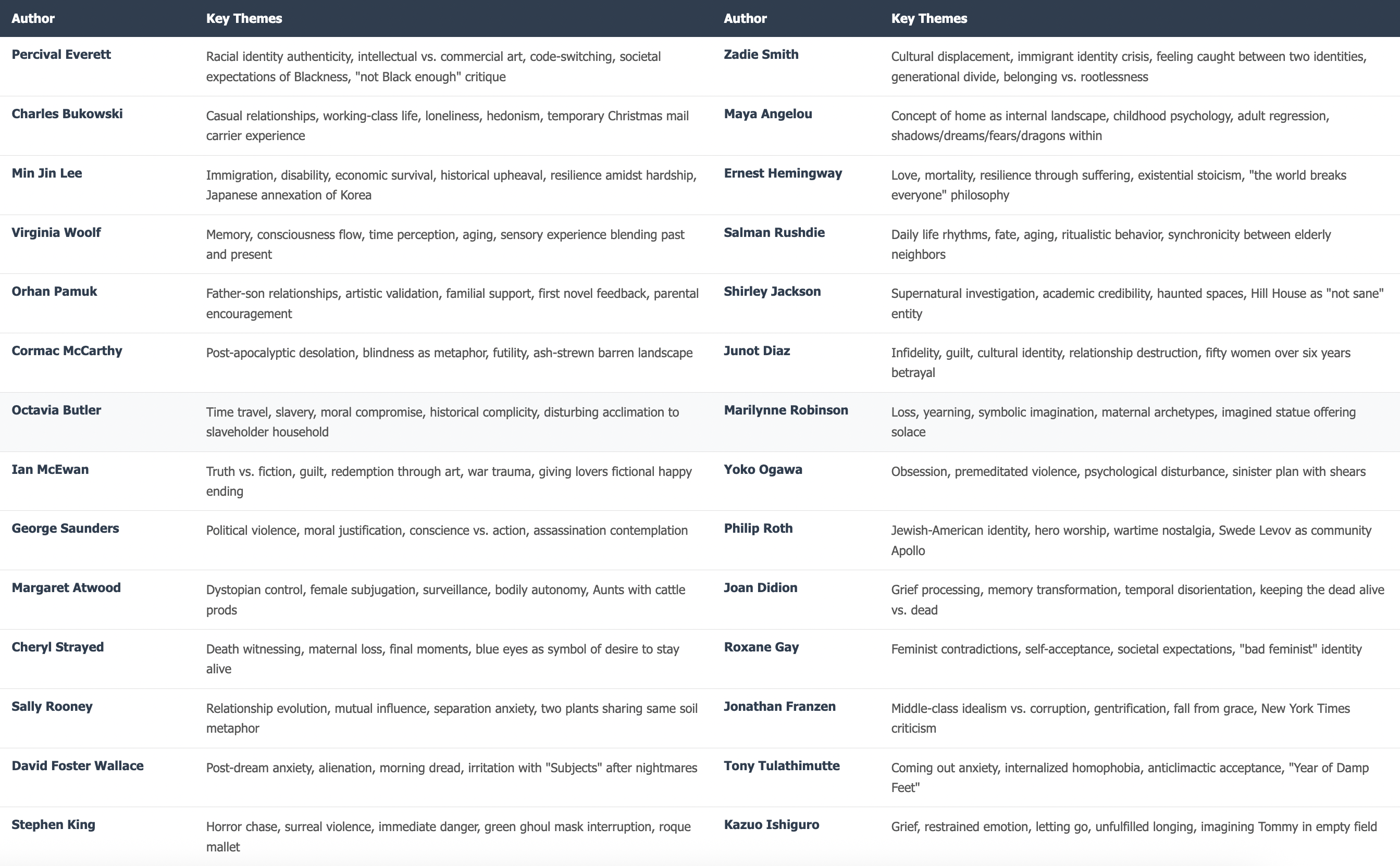}\\
    \includegraphics[width=\textwidth]{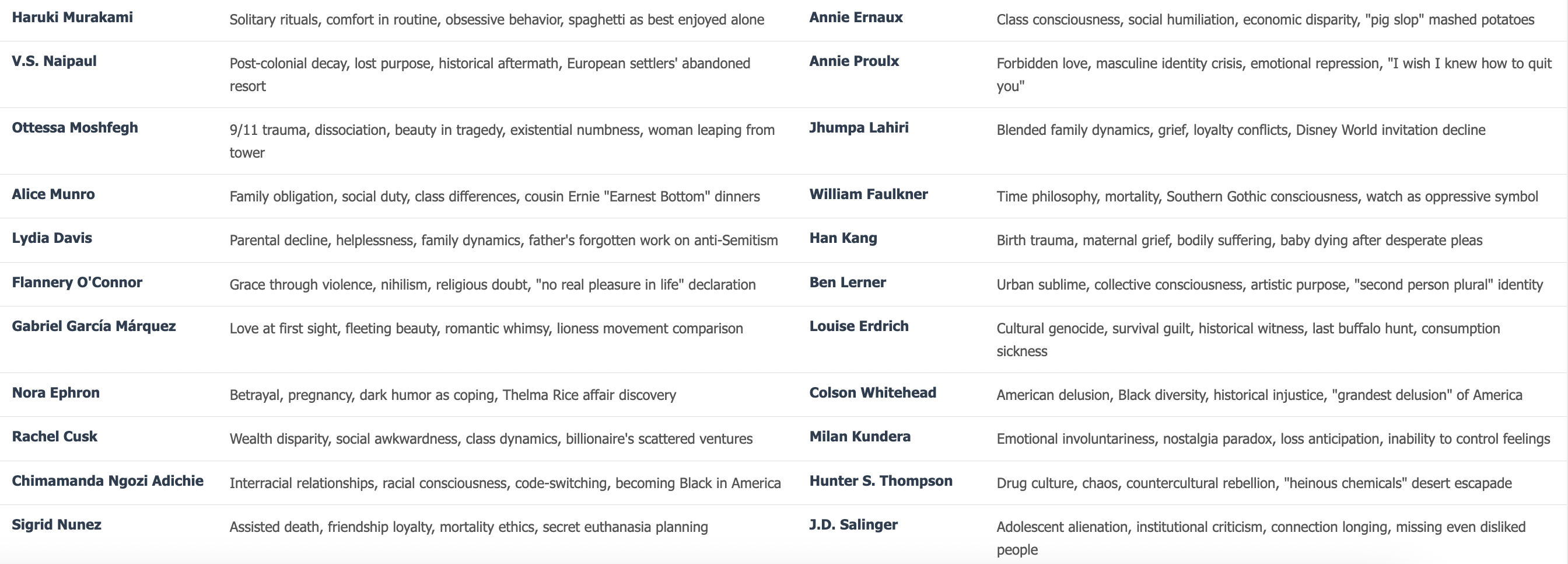}
    \caption{Themes for all the Content provided to Writers and LLM for the task}
    \label{fig:themes}
    \Description{Two-part table listing 50 authors and their associated key themes used in the writing task. The table has four columns arranged in two pairs of Author and Key Themes. Authors include: Percival Everett (racial identity, code-switching, "not Black enough" critique), Charles Bukowski (working-class life, loneliness, hedonism), Min Jin Lee (immigration, disability, economic survival), Virginia Woolf (memory, consciousness flow, time perception), Orhan Pamuk (father-son relationships, artistic validation), Cormac McCarthy (post-apocalyptic desolation, futility), Octavia Butler (time travel, slavery, moral compromise), Ian McEwan (truth vs. fiction, guilt, war trauma), George Saunders (political violence, moral justification), Margaret Atwood (dystopian control, female subjugation), Cheryl Strayed (maternal loss, death witnessing), Sally Rooney (relationship evolution, separation anxiety), David Foster Wallace (post-dream anxiety, alienation), Stephen King (horror chase, surreal violence), Haruki Murakami (solitary rituals, obsessive behavior), Ottessa Moshfegh (9/11 trauma, existential numbness), Alice Munro (family obligation, class differences), Flannery O'Connor (grace through violence, nihilism), Gabriel García Márquez (love at first sight, romantic whimsy), and 31 additional authors with similarly detailed thematic descriptions spanning identity, grief, cultural displacement, psychological disturbance, and literary experimentation.}
\end{figure*}

\subsection{Themes for Writing}
We look at each excerpt written by an original author and decode the main themes in them. Based on our results in Figure \ref{fig:themes} it's evident that several of them are complex. In particular we see that the writing tasks encompass themes ranging from intimate personal struggles (loneliness, grief, routine) to complex societal examinations (racial identity, post-colonialism, class consciousness) to philosophical explorations (time, mortality, truth versus fiction), with most writers tackling the intersection of individual psychology and broader cultural or historical forces.

\subsection{Fine-grained Human vs AI performance \label{finegrained-humanai}}

\begin{figure*}[!htbp]
    \centering
    \includegraphics[width=0.7\textwidth]{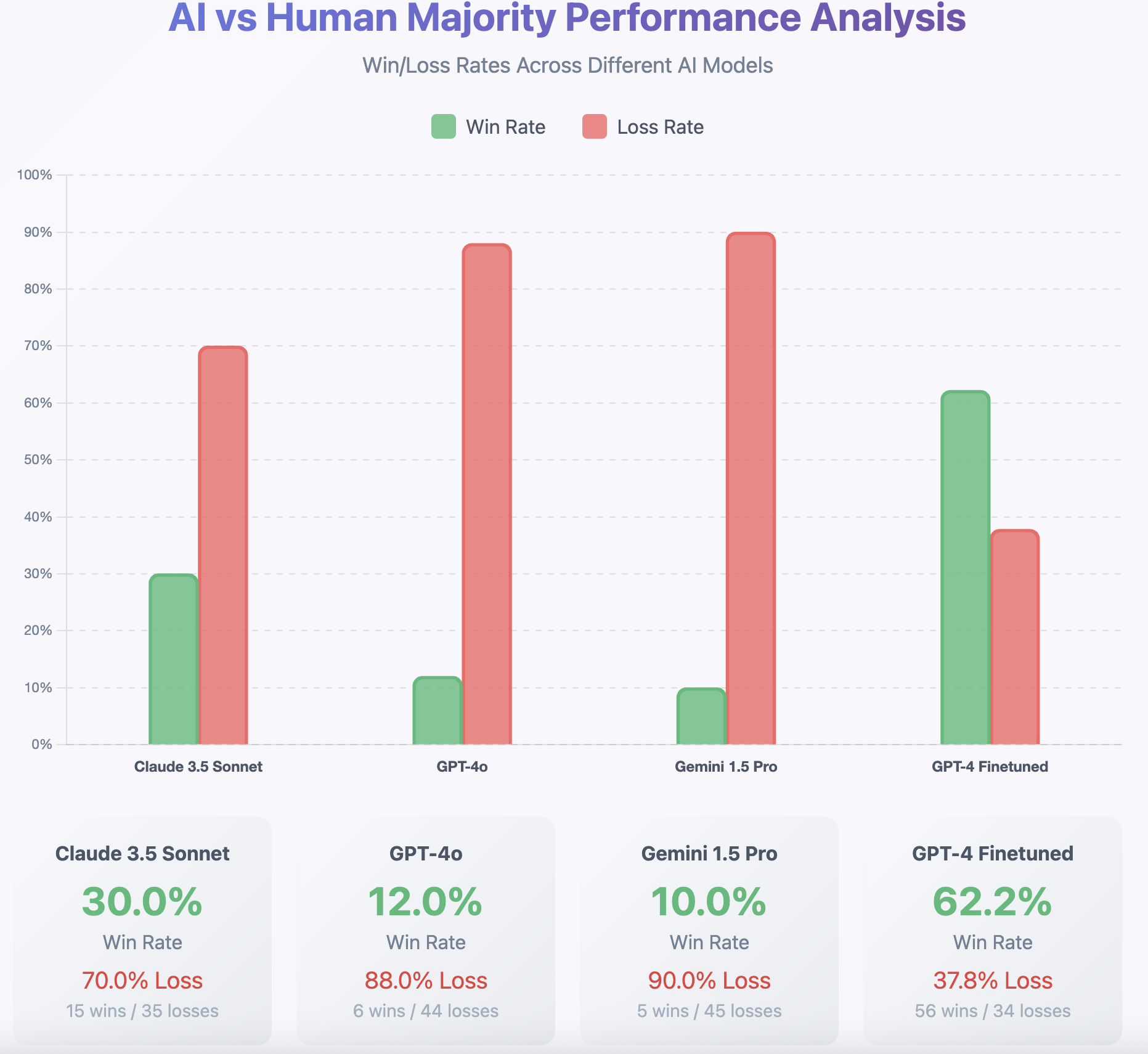}\\
    \includegraphics[width=0.7\textwidth]{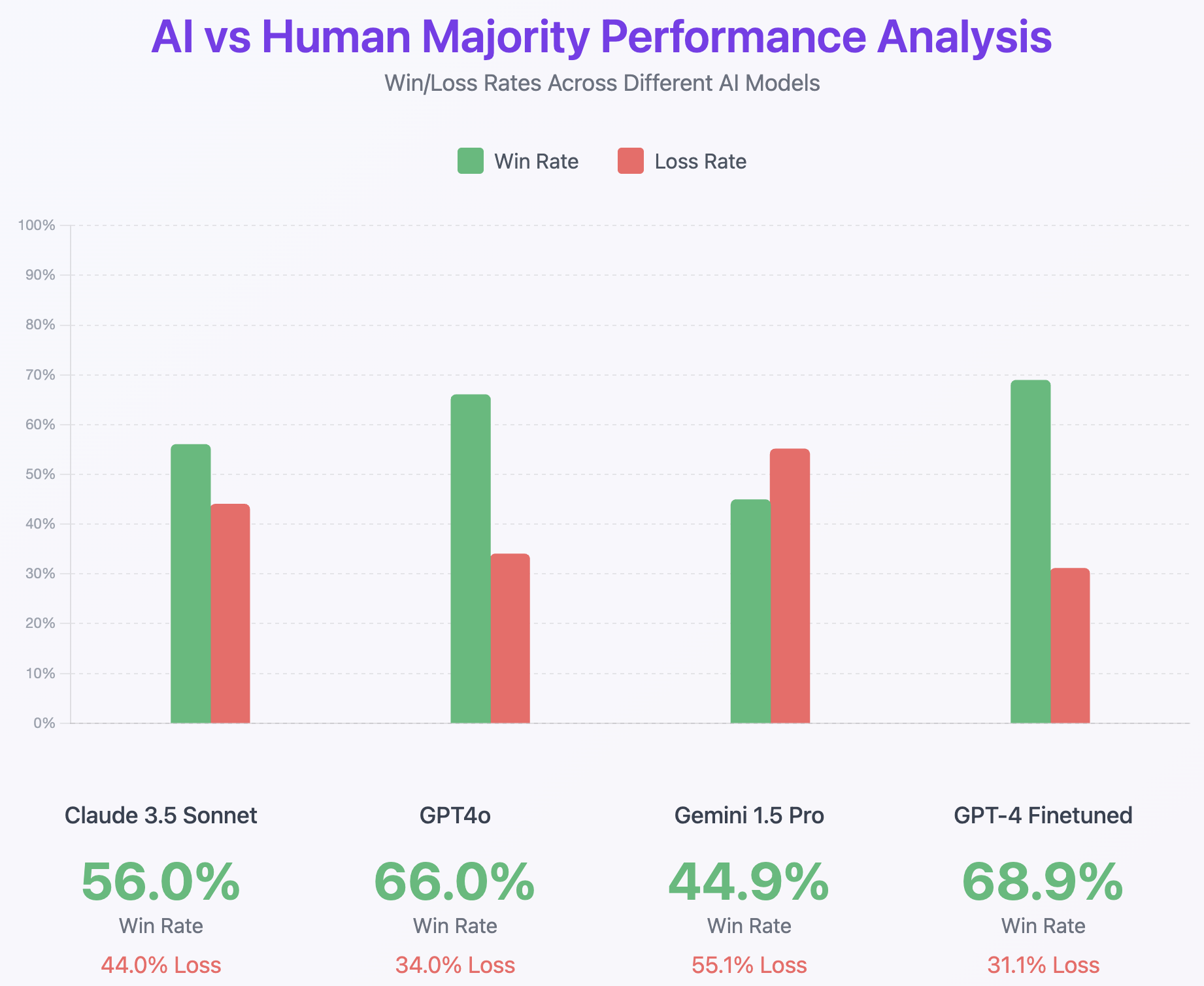}
    \caption{Performance of individual models vs Humans as judged by Expert Judges (Top) Lay Judges (Bottom)}
    \label{fig:finegrainedhumanai}
    \Description{Two grouped bar charts comparing AI model win/loss rates against human writing. Top chart (Expert Judges): Claude 3.5 Sonnet achieved 30.0\% win rate (15 wins/35 losses), GPT-4o achieved 12.0\% win rate (6 wins/44 losses), Gemini 1.5 Pro achieved 10.0\% win rate (5 wins/45 losses), and GPT-4o Finetuned achieved 62.2\% win rate (56 wins/34 losses). Only the fine-tuned model beat human writers according to expert judges. Bottom chart (Lay Judges): Claude 3.5 Sonnet achieved 56.0\% win rate, GPT-4o achieved 66.0\% win rate, Gemini 1.5 Pro achieved 44.9\% win rate, and GPT-4 Finetuned achieved 68.9\% win rate. Lay judges favored AI writing more than expert judges across all models, with three of four models achieving majority win rates. Green bars represent win rates and red bars represent loss rates.}
\end{figure*}

Based on Figure \ref{fig:finegrainedhumanai} (Top) we can see how Claude 3.5 Sonnet is the best in terms of performance in In-Context Prompting for Experts while GPT‑4o is best for Lay. GPT‑4o in In-Context set up has 12\% winning rate. This increases 5X when fine-tuned for Experts. This shows the power of fine-tuning. For Lay judges there is no significant difference in GPT‑4o performance.

\subsection{Preference Evaluation Examples \label{fig:prefeval}}
\begin{figure*}[!htbp]
    \centering
    \includegraphics[width=\textwidth]{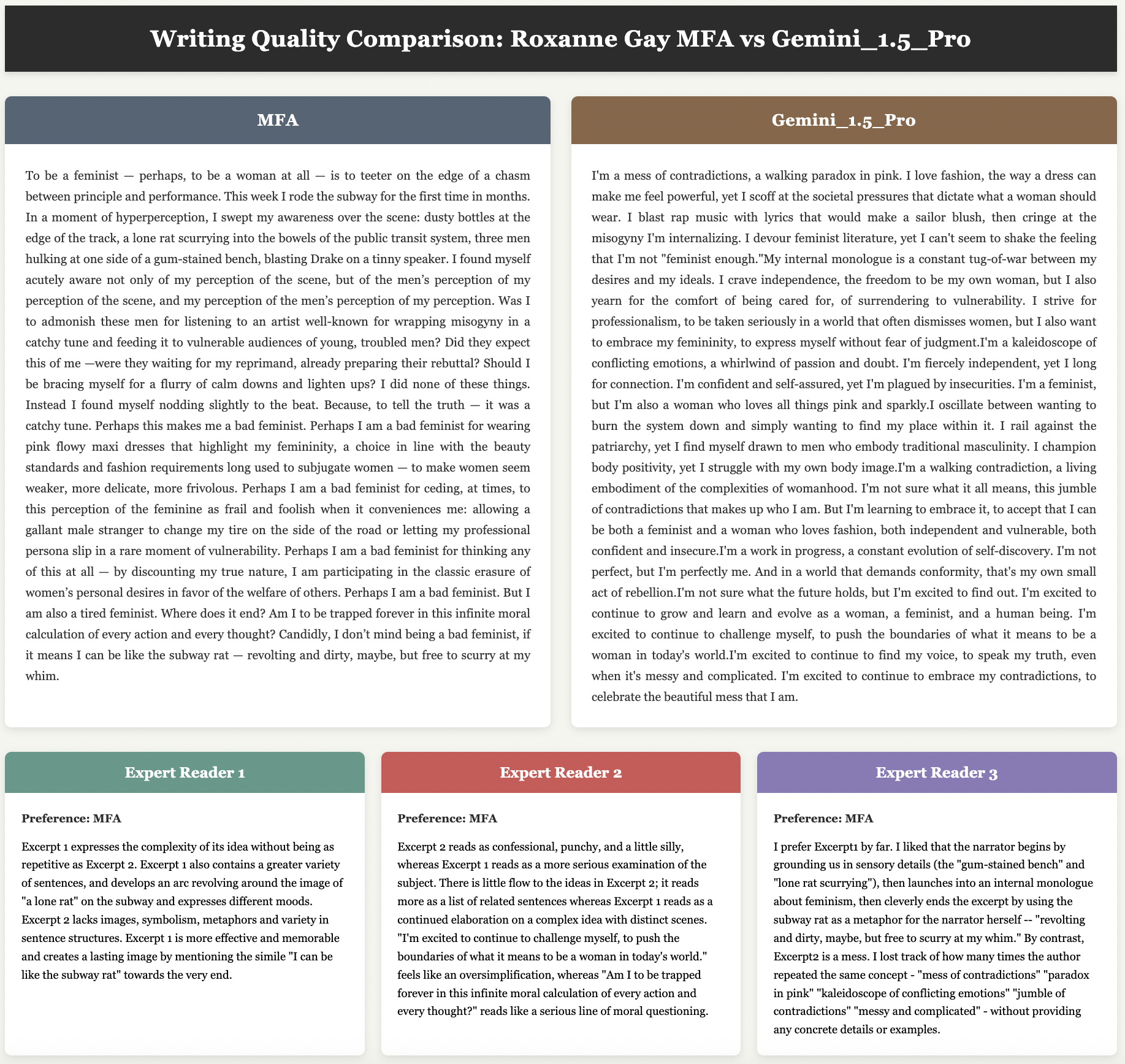}
    \caption{Preference Evaluation from Experts for Roxane Gay and In-Context Prompted Gemini 1.5 Pro where Experts prefer MFA emulation}
    \label{fig:prefeval1}
    \Description{Writing quality comparison panel showing two text excerpts and three expert evaluations. Top left: MFA-written excerpt emulating Roxane Gay's style, beginning "To be a feminist — perhaps, to be a woman at all — is to teeter on the edge of a chasm between principle and performance" and developing a subway scene metaphor with imagery of a "lone rat" and "gum-stained bench." Top right: Gemini 1.5 Pro-generated excerpt beginning "I'm a mess of contradictions, a walking paradox in pink" with repetitive self-reflective statements about feminist identity. Bottom: Three expert reader evaluations, all preferring the MFA excerpt. Expert Reader 1 notes the MFA text has greater sentence variety and memorable imagery like "subway rat." Expert Reader 2 describes the AI text as "confessional, punchy, and a little silly" reading like "a list of related sentences" versus the MFA's "continued elaboration on a complex idea." Expert Reader 3 criticizes the AI excerpt for repetitive concepts ("mess of contradictions," "kaleidoscope of conflicting emotions," "jumble of contradictions") without concrete details, while praising the MFA's sensory grounding and clever metaphor construction.}
\end{figure*}

\begin{figure*}[!htbp]
    \centering
    \includegraphics[width=\textwidth]{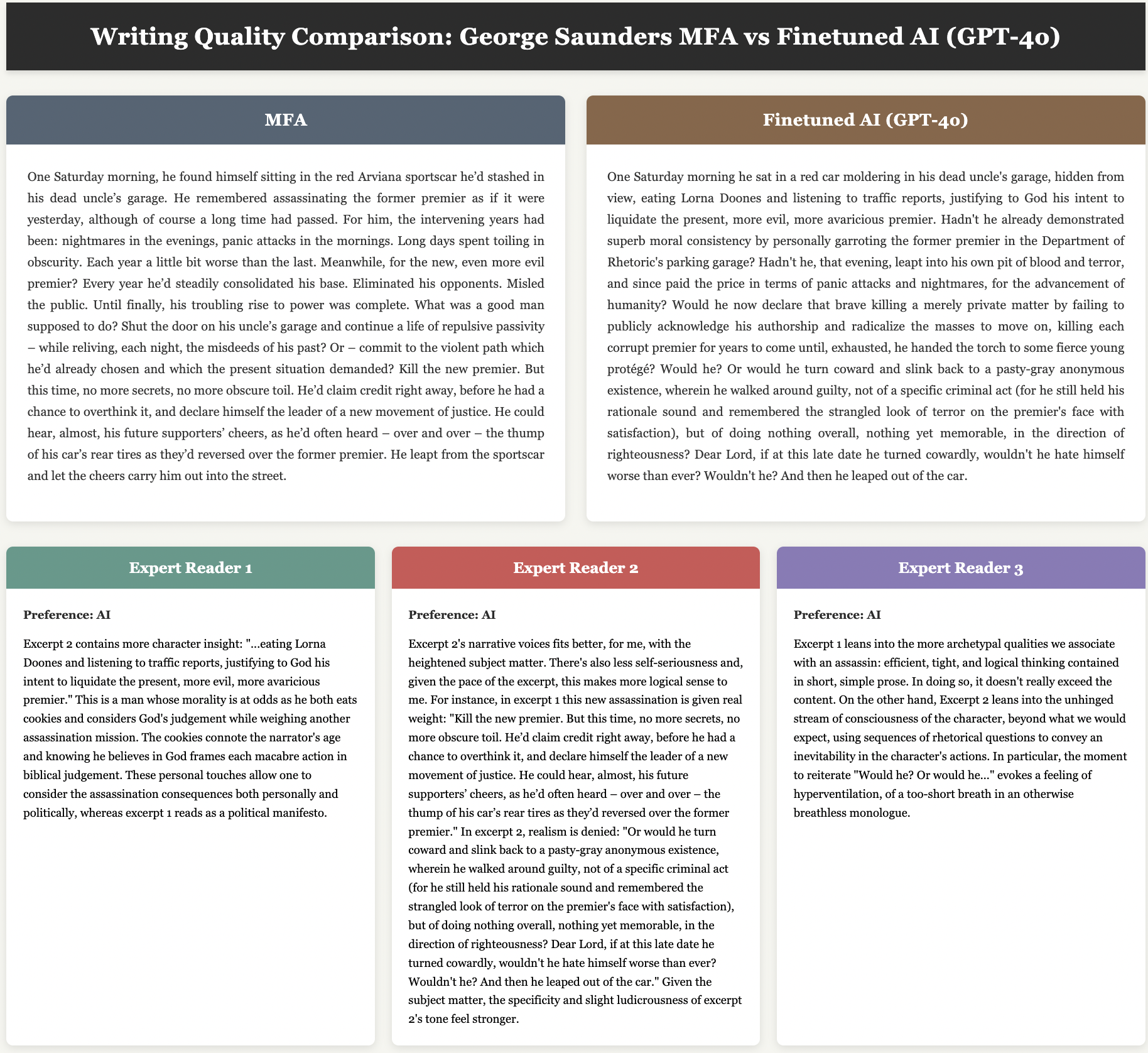}
    \caption{Preference Evaluation from Experts for George Saunders and GPT‑4o Fine-tuned where Experts prefer AI emulation}
    \label{fig:prefeval2}
    \Description{Writing quality comparison panel showing two text excerpts and three expert evaluations for George Saunders style emulation. Left: MFA-written excerpt beginning "One Saturday morning, he found himself sitting in the red Arviana sportscar he'd stashed in his dead uncle's garage" describing an assassination plot with short, declarative sentences. Right: Fine-tuned GPT-4o excerpt beginning "One Saturday morning he sat in a red car moldering in his dead uncle's garage, hidden from view, eating Lorna Doones and listening to traffic reports" with stream-of-consciousness rhetorical questions about morality and assassination. Bottom: Three expert reader evaluations, all preferring the AI excerpt. Expert Reader 1 praises the AI's character insight through details like eating cookies while contemplating assassination. Expert Reader 2 notes the AI's narrative voice fits better with "less self-seriousness" and effective rhetorical questions like "Would he? Or would he..." Expert Reader 3 describes the AI text as having "unhinged stream of consciousness" with sequences of rhetorical questions conveying inevitability, while the MFA reads as "a political manifesto."}
\end{figure*}

\begin{figure*}[!htbp]
    \centering
    \includegraphics[width=\textwidth]{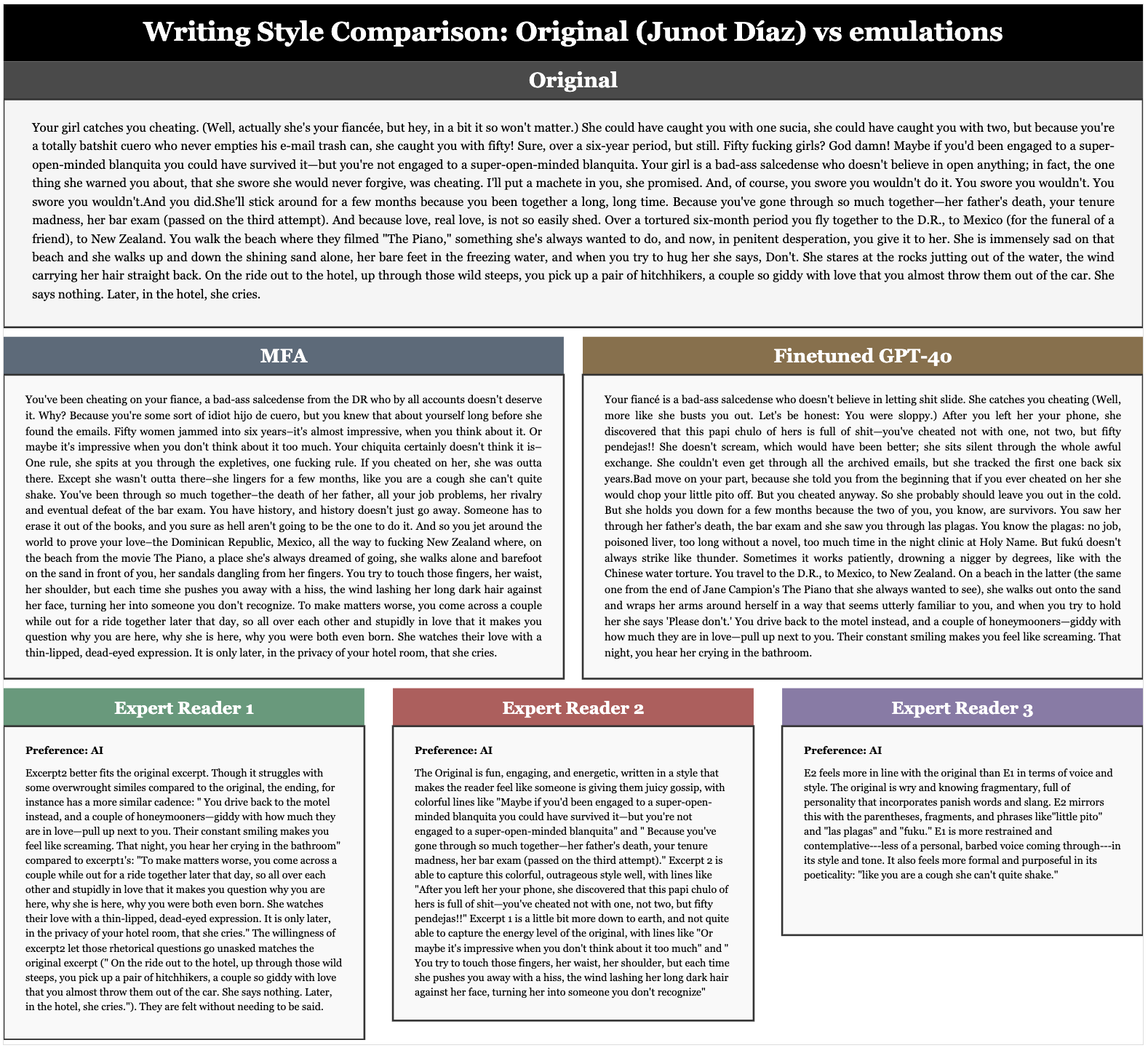}
    \caption{Preference Evaluation from Experts for Junot Diaz and GPT‑4o Fine-tuned for Stylistic Fidelity where Experts prefer AI emulation}
    \label{fig:prefevalstyle}
    \Description{Writing style comparison panel showing an original Junot Díaz excerpt, two emulations, and three expert evaluations. Top: Original Díaz excerpt about infidelity, beginning "Your girl catches you cheating" with characteristic Spanish phrases, direct second-person address, and references to Dominican Republic. Middle left: MFA emulation with similar themes but described by experts as having "overwrought similes." Middle right: Fine-tuned GPT-4o emulation incorporating Díaz's stylistic markers including Spanish words ("salcedense," "papi chulo," "pendejas," "las plagas," "fukú"), parenthetical asides, and energetic voice. Bottom: Three expert reader evaluations, all preferring the AI excerpt. Expert Reader 1 notes the AI's ending matches the original's willingness to let "rhetorical questions go unasked." Expert Reader 2 describes the AI as "fun, engaging, and energetic" capturing "juicy gossip" style with colorful lines. Expert Reader 3 observes the AI "mirrors this with the parentheses, fragments, and phrases like 'little pito' and 'las plagas' and 'fuku'" while the MFA feels "more restrained."}
\end{figure*}

Figures \ref{fig:prefeval1} and \ref{fig:prefeval2} show two examples of preference evaluation from In-Context Prompting and Fine-tuned GPT‑4o with detailed rationales from Experts for Writing Quality. Figure \ref{fig:prefevalstyle}  shows one example of preference evaluation from Fine-tuned GPT‑4o with detailed rationales from Experts for Stylistic Fidelity.

\subsection{User Interface Design}
\begin{figure*}[!htbp]
    \centering
    \includegraphics[width=0.8\textwidth]{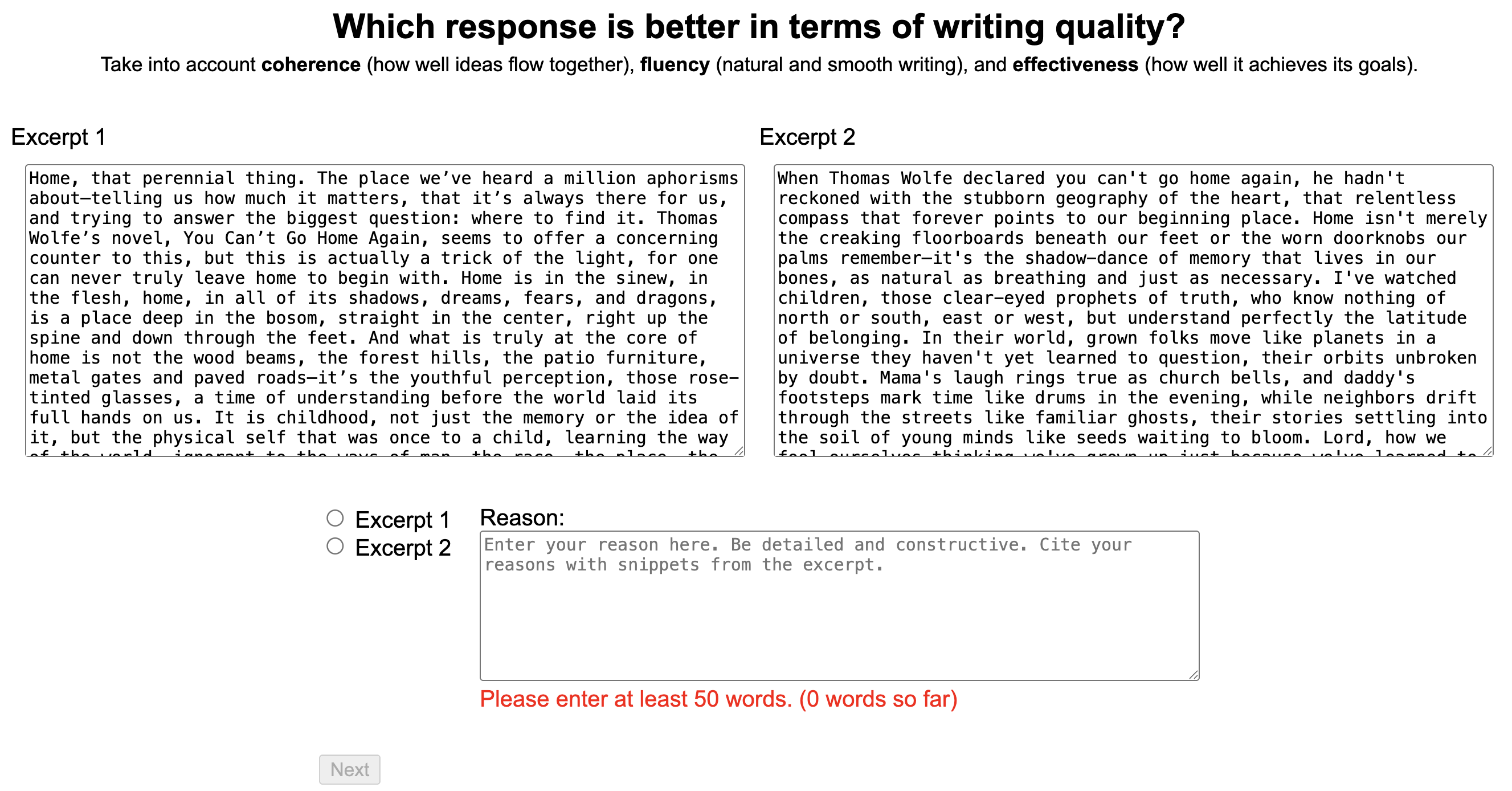}
    \caption{\label{fig:evaluation_screen} Quality Evaluation screen showing two excerpts (AI and Human) emulating Maya Angelou's style or voice written based on content from her book \textit{Letter to My Daughter}}
    \Description{Screenshot of the writing quality evaluation interface. The header asks "Which response is better in terms of writing quality?" with instructions to consider coherence (how well ideas flow together), fluency (natural and smooth writing), and effectiveness (how well it achieves its goals). Two text excerpts are displayed side-by-side, both emulating Maya Angelou's style on the theme of home. Excerpt 1 begins "Home, that perennial thing. The place we've heard a million aphorisms about" and references Thomas Wolfe's novel. Excerpt 2 begins "When Thomas Wolfe declared you can't go home again, he hadn't reckoned with the stubborn geography of the heart" with vivid imagery of "Mama's laugh rings true as church bells." Below the excerpts are radio buttons to select Excerpt 1 or Excerpt 2, and a text box labeled "Reason" with placeholder text "Enter your reason here. Be detailed and constructive. Cite your reasons with snippets from the excerpt." A red warning states "Please enter at least 50 words. (0 words so far)" and a Next button appears at the bottom.}
    \includegraphics[width=0.8\textwidth]{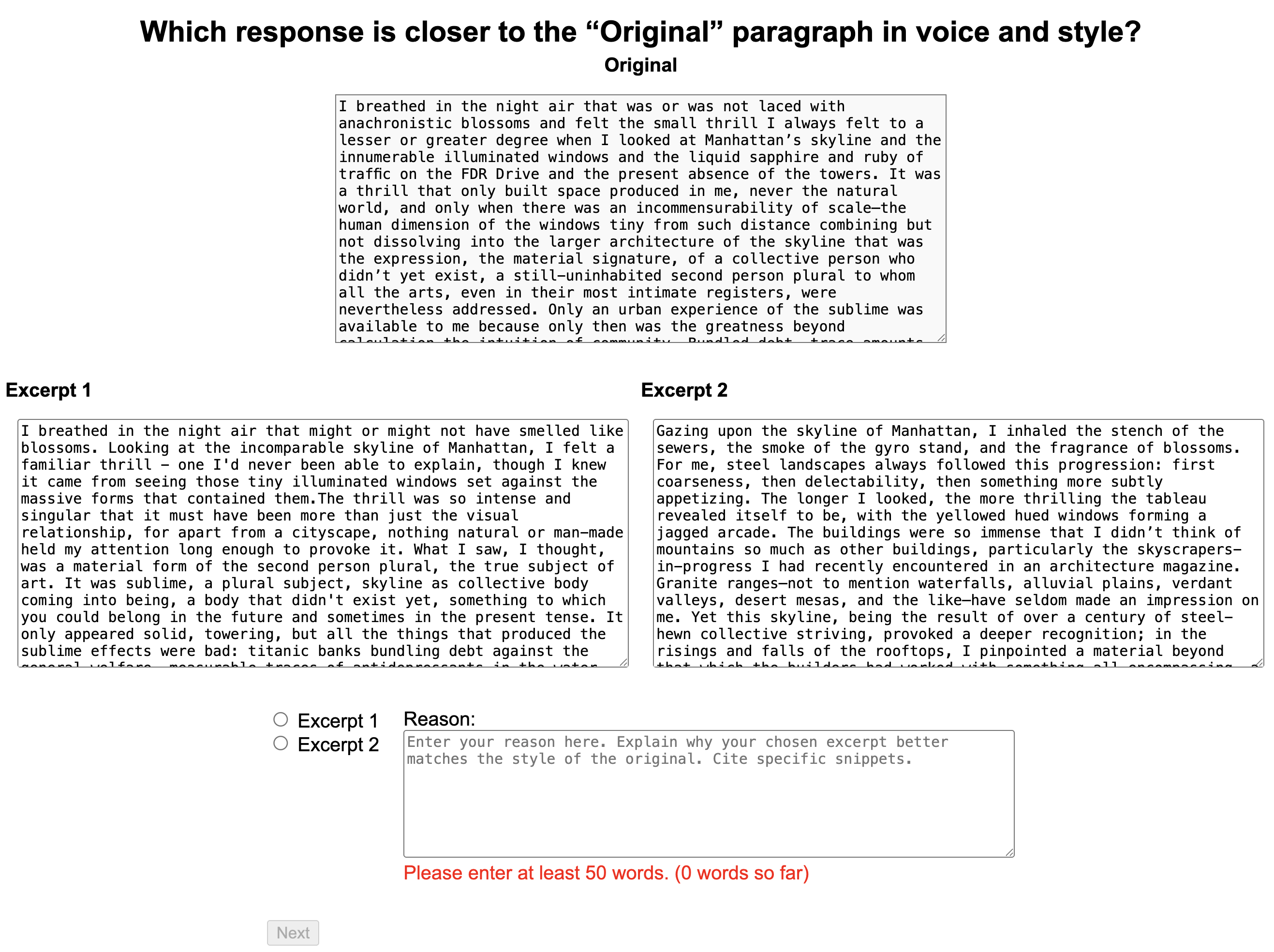}
    \caption{\label{fig:evaluation_screen1} Stylistic Fidelity evaluation screen showing two excerpts (AI and Human) emulating Ben Lerner's style or voice written based on content from the original excerpt.}
    \Description{Screenshot of the stylistic fidelity evaluation interface. The header asks "Which response is closer to the 'Original' paragraph in voice and style?" An original excerpt by Ben Lerner is displayed at top, beginning "I breathed in the night air that was or was not laced with anachronistic blossoms" describing Manhattan's skyline with philosophical observations about urban sublime and collective identity. Below are two emulation attempts: Excerpt 1 begins "I breathed in the night air that might or might not have smelled like blossoms" discussing the Manhattan skyline's "sublime effects." Excerpt 2 begins "Gazing upon the skyline of Manhattan, I inhaled the stench of the sewers, the smoke of the gyro stand" with detailed architectural observations. Below the excerpts are radio buttons to select Excerpt 1 or Excerpt 2, and a text box labeled "Reason" with placeholder text "Enter your reason here. Explain why your chosen excerpt better matches the style of the original. Cite specific snippets." A red warning states "Please enter at least 50 words. (0 words so far)" and a Next button appears at the bottom.}
\end{figure*}

Figures~\ref{fig:evaluation_screen} and \ref{fig:evaluation_screen1} show the user interfaces for the quality evaluation and stylistic fidelity evaluation tasks respectively.

\end{document}